\documentclass[lettersize,journal]{IEEEtran}
\usepackage{amsmath,amsfonts}
\usepackage{algorithmic}
\usepackage{algorithm}
\usepackage{array}
\usepackage[caption=false,font=normalsize,labelfont=sf,textfont=sf]{subfig}
\usepackage{textcomp}
\usepackage{stfloats}
\usepackage{cite}
\usepackage{hyperref}
\usepackage{url}
\usepackage{graphicx}
\usepackage[numbers]{natbib}
\usepackage{enumitem}
\usepackage{booktabs}
\usepackage{multirow}
\usepackage{geometry}
\usepackage[utf8]{inputenc}
\usepackage{array}
\usepackage{xcolor}
\usepackage{colortbl}
\usepackage{fancyvrb}
\usepackage{mdframed}
\usepackage{multicol}
\usepackage{verbatim}
\usepackage{listings}
\usepackage{subcaption}
\usepackage{hyperref}
\definecolor{posgreen}{RGB}{0,128,0}
\definecolor{negred}{RGB}{220,20,60}
\definecolor{tealblue}{RGB}{102, 194, 164}
\definecolor{lightbg}{RGB}{248, 252, 251}
\definecolor{headercolor}{RGB}{0,0,0}
\definecolor{origincolor}{RGB}{200,200,200}
\definecolor{trainedcolor}{RGB}{220,220,220}

\newcommand{\coloracc}[1]{%
  \ifdim#1pt>0pt
    \textcolor{posgreen}{#1}%
  \else
    \ifdim#1pt<0pt
      \textcolor{negred}{#1}%
    \else
      #1%
    \fi
  \fi
}

\newcommand{\colorlen}[1]{%
  \ifdim#1pt<0pt
    \textcolor{posgreen}{#1}%
  \else
    \ifdim#1pt>0pt
      \textcolor{negred}{#1}%
    \else
      #1%
    \fi
  \fi
}

\newcommand{\orcid}[1]{\href{https://orcid.org/#1}{\includegraphics[height=8pt]{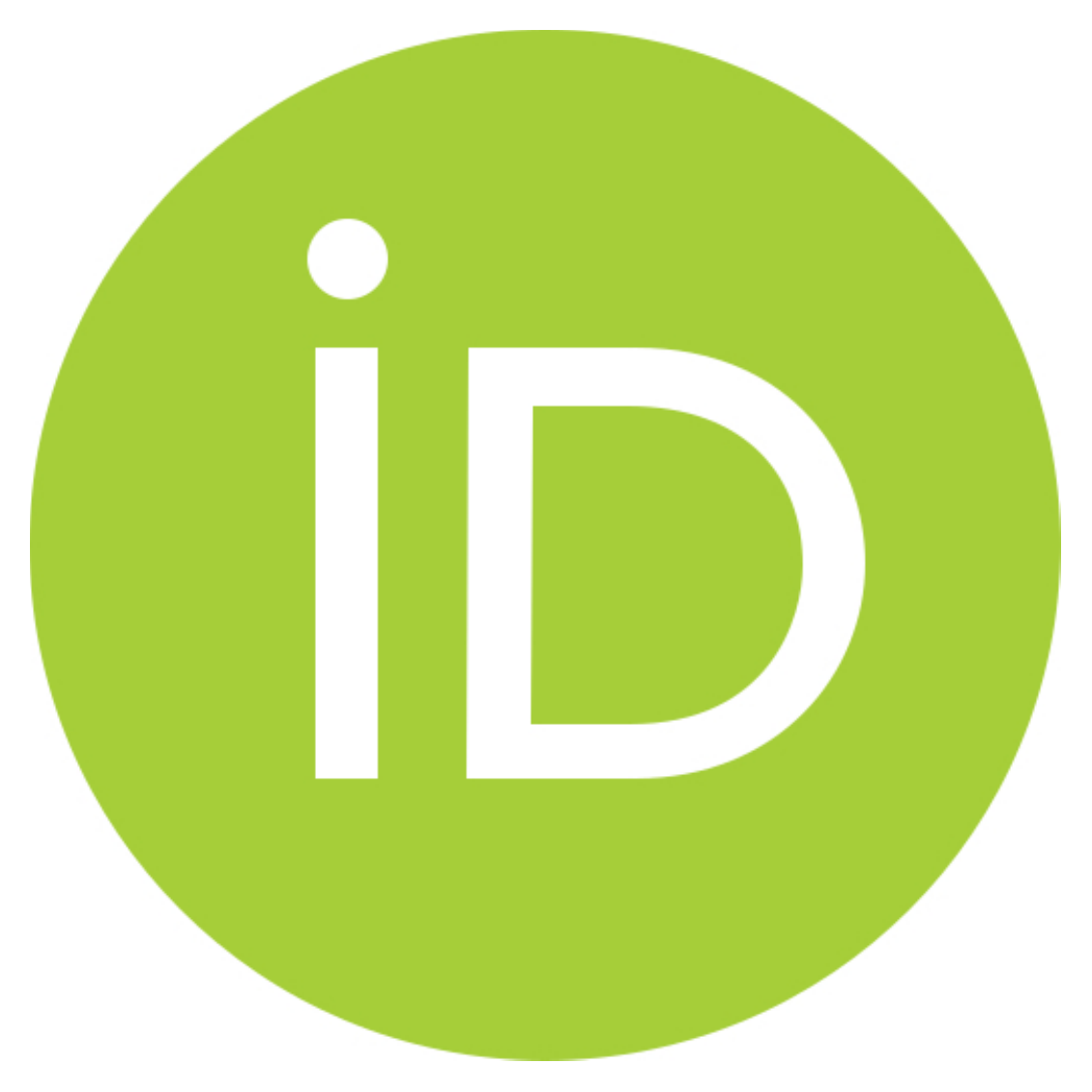}}}

\begin{document}

\title{Explore Briefly, Then Decide: Mitigating LLM Overthinking via Cumulative Entropy Regulation}

\author{
Yi Bin\orcid{0000-0001-9714-8738}, Tianyi Jiang\orcid{0009-0000-4039-9982}, Yujuan Ding\orcid{0000-0003-2945-1107}, Kainian Zhu\orcid{0009-0004-9631-7228}, Fei Ma\orcid{0009-0002-5388-9125}, Jingkuan Song\orcid{0000-0002-2549-8322}~\IEEEmembership{Senior Member,~IEEE}, Yang Yang\orcid{0000-0002-5070-4511}~\IEEEmembership{Senior Member,~IEEE}, and Heng Tao Shen\orcid{0000-0002-2999-2088}~\IEEEmembership{Fellow,~IEEE}

\thanks{Corresponding author: Heng Tao Shen,  shenhengtao@hotmail.com}
\thanks{Tianyi Jiang, Yi Bin, Jingkuan Song and Heng Tao Shen is with Tongji University, Shanghai, China.}
\thanks{Yujuan Ding is with Hong Kong Polytechnic University, Hong Kong SAR, China.}
\thanks{Kainian Zhu is with Shanghai University of Electric and Power, Shanghai, China.}
\thanks{Fei Ma is with Guangdong Laboratory of Artificial Intelligence and Digital Economy, Guangdong, China.}
\thanks{Yang Yang is with University of Electronic Science and Technology of China, Chengdu, China.}
}


\markboth{Journal of \LaTeX\ Class Files,~Vol.~14, No.~8, August~2021}%
{Jiang \MakeLowercase{\textit{et al.}}: Explore Briefly, Then Decide: Mitigating LLM Overthinking via Cumulative Entropy Regulations}

\IEEEpubid{0000--0000/00\$00.00~\copyright~2021 IEEE}

\maketitle

\begin{abstract}
Large Language Models (LLMs) have demonstrated remarkable reasoning abilities on complex problems using long Chain-of-Thought (CoT) reasoning. However, they often suffer from overthinking, meaning generating unnecessarily lengthy reasoning steps for simpler problems. This issue may degrade the efficiency of the models and make them difficult to adapt the reasoning depth to the complexity of problems. To address this, we introduce a novel metric \textbf{T}oken \textbf{E}ntropy \textbf{C}umulative \textbf{A}verage (\textbf{TECA}), which measures the extent of exploration throughout the reasoning process. We further propose a novel reasoning paradigm named “Explore Briefly, Then Decide”, with an associated \textbf{C}umulative \textbf{E}ntropy \textbf{R}egulation (\textbf{CER}) mechanism. This paradigm leverages TECA to help the model dynamically determine the optimal point to conclude its thought process and provide a final answer, thus achieving efficient reasoning. Experimental results across diverse mathematical benchmarks show that our approach substantially mitigates overthinking without sacrificing problem-solving ability. With our thinking paradigm, the average response length decreases by up to 71\% on simpler datasets, demonstrating the effectiveness of our method in creating a more efficient and adaptive reasoning process. Code is available in this \href{https://github.com/AusertDream/CumulativeEntropyRegulation}{http URL}.
\end{abstract}


\section{Introduction}
\IEEEPARstart{L}{arge} Language Models (LLMs) have demonstrated remarkable capabilities in complex problem-solving, particularly when using the Chain-of-Thought (CoT) mechanism. This method breaks down difficult problems into a series of intermediate steps, known as the “thinking” process, enhancing the model's reasoning ability. Further, methods like GSPO, DAPO, and Multi-layer GRPO~\cite{zhengGroupSequencePolicy2025,yuDAPOOpenSourceLLM2025,dingMultiLayerGRPOEnhancing2025} have been developed specifically to help models tackle increasingly difficult problems with Long CoT, while may raise another significant challenge termed as \textbf{overthinking}.

\begin{figure*}
    \centering
    \includegraphics[width=0.9\linewidth]{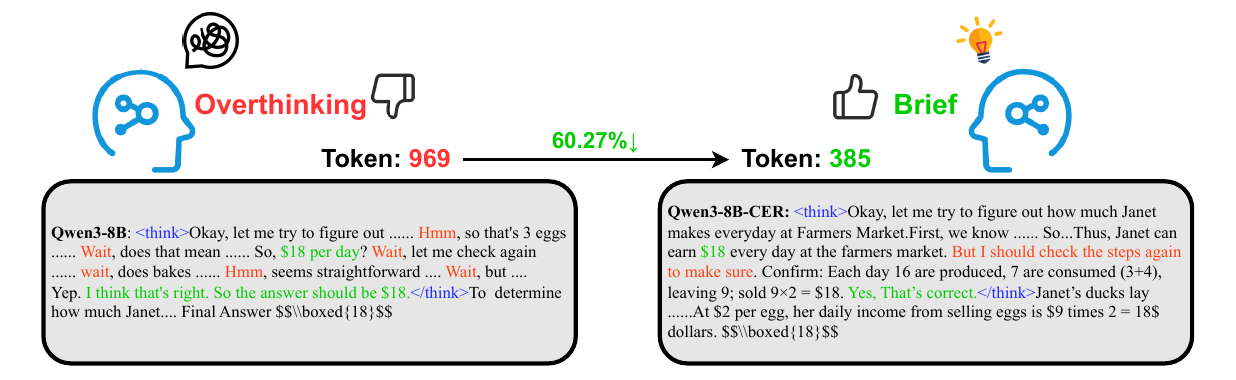}
    \caption{Reasoning process comparison between the original and CER-trained large reasoning models.
    The original model continues to reflect for four times after the correct answer appears, while the CER-trained model determines the final answer after only one reflection.
    GREEN: correct answers (first and last shown); RED: reflecting words. }
    \label{fig0:case_illustration}
\end{figure*}

Overthinking means generate unnecessarily long and redundant reasoning steps, even for simple problems~\cite{cuesta-overthinkingEvidence2025}, as an example shown in Figure~\ref{fig0:case_illustration}. This not only increases computational costs but can also degrade reasoning accuracy, as models may disregard a correct solution they've already found in favor of further, often incorrect, exploration. This issue makes it difficult for LLMs conduct efficient inference, or adapt the depth of their reasoning to the complexity of the task, therefore has attracted considerable research attention~\cite{suiStopOverthinkingSurvey2025, shi2025multimodal, yue2025dontoverthinkitsurvey,xuCoD2025,nayabCCoT2025,huangCGRS2025,jinReCUTBalancingReasoning2025,hongReconsideringOverthinkingPenalizing2025,yang2025deeRM}. 
\IEEEpubidadjcol
Reinforcement Learning (RL) is one of the effective approaches to train a reasoning model for better capability(e.g., DeepSeek-R1~\cite{deepseekr1}, DeepSeek-R1-Zero~\cite{deepseekr1}, OpenAI o1~\cite{openaiOpenAIO1System2024}, QwQ-32B-Preview~\cite{qwq32b, qwen2.5}), while generally the learning focuses on the accuracy reward and format rewards. To tackle the overthinking issue, it is also natural to think about integrating thinking length-related reward into the RL framework~\cite{yue2025dontoverthinkitsurvey}. For example, the thinking length reward assigns higher scores to short, correct solutions while penalizing lengthy or incorrect ones, thereby optimizing the length of the reasoning path. Several existing methods have made various attempts, for example to apply CoT steps, Token numbers or other metrics to build the reward~\cite{suiStopOverthinkingSurvey2025,aggarwal2025l1,arora2025training}. However, even though simply constraining the CoT length or token count can reduce the output length, they do not necessarily optimize the underlying thinking process. Instead, it may harm essential reasoning steps, forcing the model to prematurely cut short a necessary thought process, which could lead to an incorrect final answer. This creates a trade-off between length and accuracy.

To overcome this limitation, we conduct the analysis of the reasoning process, and summarize that the process generally contains two distinct stages: \textbf{exploration} and \textbf{determination}. The exploration stage involves generating new paths and ideas~\cite{wang8020Rule2025}, which is typically marked by high token entropy (representing the model's high uncertainty and the wide range of possible next tokens). The determination stage, on the other hand, is when the model follows a single, established path~\cite{wang8020Rule2025}, with low token entropy. Based on this, we hypothesize that overthinking is caused by unnecessary over-exploration. Building on this insight, we propose an effective metric to indicate the overthinking stage: Token Entropy Cumulative Average (TECA). TECA is a simple cumulative average~\cite{kiliCumulativeAverage} of all token entropy up untill the current point~\cite{wang8020Rule2025} that represents the model's uncertainty at each inference step. By tracking TECA, we can directly identify the existence of ``forking tokens'' and, therefore, the exploration stage~\cite{wang8020Rule2025}. Our experimental analysis confirm this hypothesis, as shown in Figure.~\ref{fig1:ATE_study}, short-thinking models\footnote{The models don't have the long CoT ability.} exhibit a TECA that increases briefly at the beginning of a reasoning process and then drops, while long CoT models show a prolonged increase, suggesting excessive and unnecessary exploration. We further propose a novel thinking paradigm: ``\textit{\textbf{Explore Briefly, Then Decide}}'', which tries to mimic efficient human reasoning, where the goal of problem-solving is to reach a decisive conclusion, not to endlessly explore alternative paths. We apply our TECA into the thinking paradigm by introducing the Cumulative Entropy Regulation (CER) to train an efficient reasoning model within a GRPO reinforcement learning algorithm. CER is able to suppress the model's excessive exploration while preserving its necessary exploration abilities. We further enhance this by using a segmented reward mechanism, applying CER only to correctly predicted samples to improve learning efficiency. 

To evaluate our method, we conducted extensive experiments on four math problem benchmarks. The results demonstrate that our approach effectively reduces the response length of models with only a small change in accuracy, as a case shown in Figure.~\ref{fig0:case_illustration}. More specifically, our method reduced the response length of Qwen3-4B by 71\% on GSM8K and 39.25\% on MATH500. For Qwen3-8B, we saw a reduction of 55.21\% on GSM8K and 32.76\% on MATH500. Our performance also consistently outperforms other existing methods for addressing overthinking. The TECA curves of our trained models visually confirm the desirable ``Explore Briefly, Then Decide'' thinking pattern, further validating the effectiveness of using TECA as a metric to adaptively adjust the thinking process. By teaching the model to briefly explore and then determine the answer, we enable it to choose the correct extent of exploration, thereby significantly reducing the occurrence of overthinking.

\section{Related work}
\subsection{Overthinking Problem.} DeepSeek R1 demonstrates that long reasoning ability of LLM can be activated by RL~\cite{deepseekr1}, however Large Reasoning Models have a severe overthinking problem which manifests as having to think a lot even about simple questions~\cite{wuWhenMoreLess2025}. For concise reasoning, there are many solutions~\cite{suiStopOverthinkingSurvey2025}. One of solutions~\cite{wangWaitWeDont2025} is that if the self-reflection tokens such as “Wait” and “Hmm” are suppressed, the response length will decrease up to 27\%–51\%. Another finding is~\cite{liaoLostBeginningReasoning2025} that the first reasoning step is crucial. Choosing a correct first reasoning step will reduce the inference cost up to 70\%. Moreover, some works start with improving the RL algorithm. Using length-dependent rewards and advantage reweighting can make response length align with the difficulty of the question~\cite{zhangGRPOLEADDifficultyAwareReinforcement2025}. GFPO~\cite{shrivastavaSampleMoreThink2025} proposes two metrics: token efficiency and response length for group filtering in GRPO which reduce the response length 46-71\% with response length and 71-85\% with token efficiency. In addition to modifying the algorithms themselves, there are also works proposing new training paradigms, such as Multi-Stage RL~\cite{tuLearningWhenThink2025}, D-CoT~\cite{wangDynamicChainofThoughtAdaptive2025}, QFFT~\cite{liuQFFTQuestionFreeFineTuning2025}, IBPO~\cite{yuThinkSmarterNot2025}. Unlike these works, our work uses TECA to analyze the process of reasoning and propose CER which will lead model to learn ``Explore Briefly, Then Decide'' thinking paradigm.

\subsection{Entropy Mechanism.} Information entropy is a concept in information theory which quantifies the average level of uncertainty or information associated with the variable's potential states or possible outcomes~\cite{entropy}. In practical applications of entropy, it is primarily divided into entropy-based optimization methods and entropy-based problem research. Entropy-based optimization add information entropy into the train metrics such as loss, object function. Gao et al. ~\cite{gaoOneshotEntropyMinimization2025} uses just one single unlabeled carefully selected data to train the model. After only 10 steps optimization, they achieve performance improvements greater than those obtained using thousands of data and carefully designed rewards in rule-based reinforcement learning. Similarly, Agarwal et al. ~\cite{agarwalUnreasonableEffectivenessEntropy2025} proposes three kinds of entropy train method: EM-FT, EM-RL and EM-INF. Especially, EM-INF enables Qwen-32B to match or exceed the performance of proprietary models like GPT-4o, Claude 3 Opus, and Gemini 1.5 Pro on the challenging SciCode benchmark which reveals that many pretrained LLMs possess previously underappreciated reasoning capabilities that can be effectively elicited through entropy minimization alone, without any labeled data or even any parameter updates. On the other hand, Wang et al. ~\cite{wang8020Rule2025} proposed the ``forking token'' concept which is high-entropy minority token during model answering and such token will lead the model to fork the path and explore more possibility. More important, if gradient updates at the forking tokens instead of all tokens which means stronger desire to explore, the trained model will get higher accuracy on AIME24 and AIME25 contrast to the trained at all tokens. It reveals that the forking tokens make more contribution to answering question correctly. Cui et al. ~\cite{cuiEntropyMechanismReinforcement2025} reveals an inverse relationship between the downstream task performance of large reasoning models trained with reinforcement learning and entropy, and proposes a solution to promote exploration and escape the entropy collapse problem. 

\begin{figure*}[t!] 
    \centering
    \subfloat[One case for short-CoT models]{
        \includegraphics[width=0.45\textwidth]{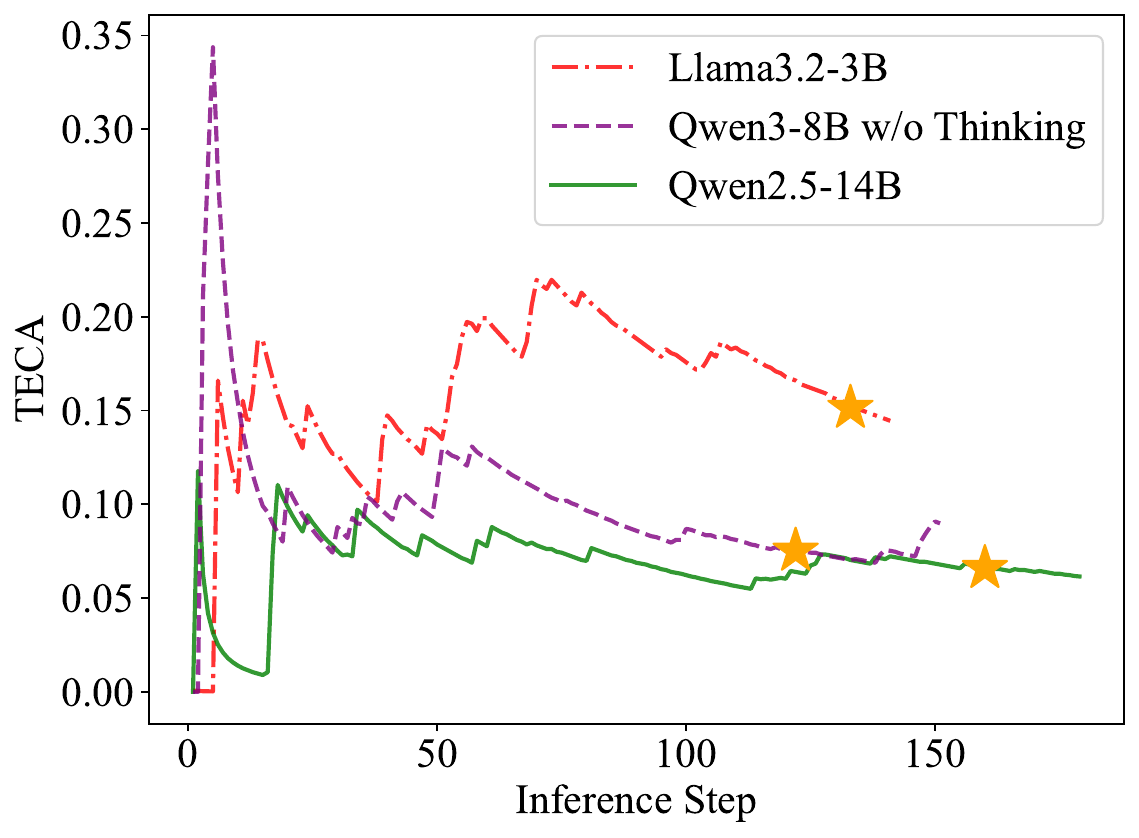}
        \label{fig1:a}
    }
    \hfil 
    \subfloat[One case for a long-CoT model]{
        \includegraphics[width=0.45\textwidth]{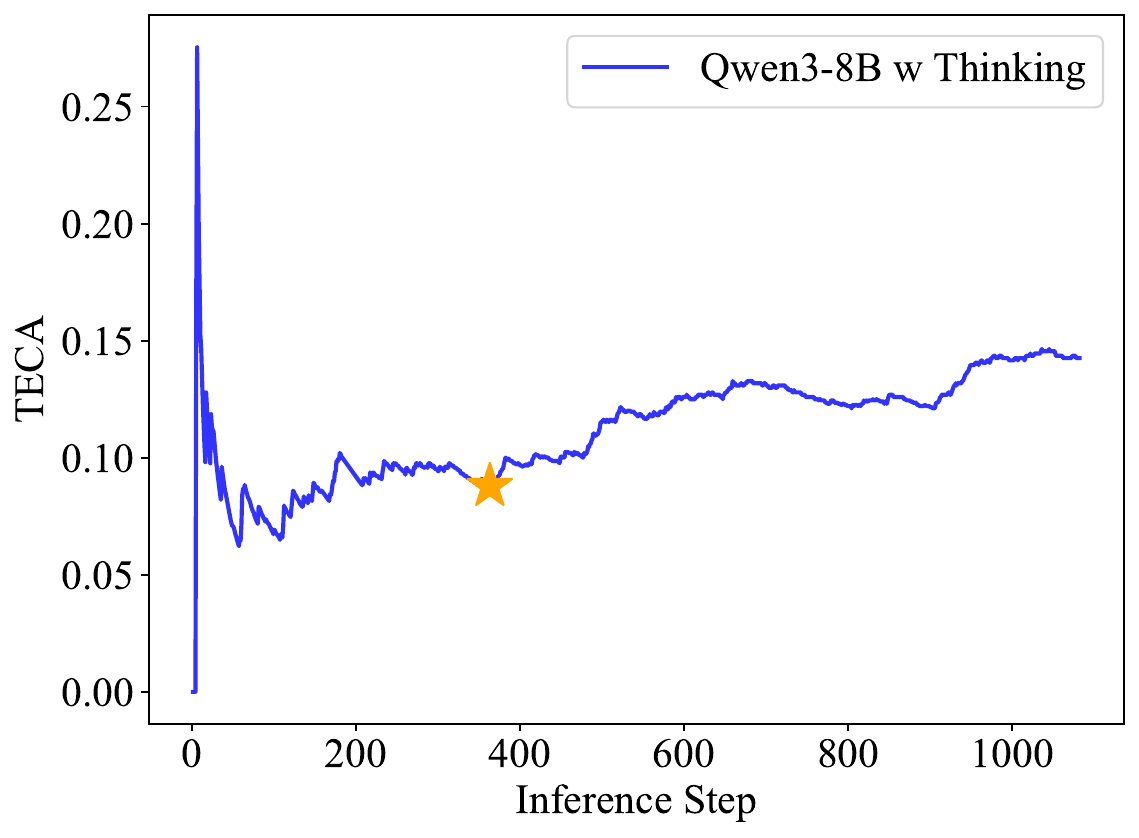}
        \label{fig1:b}
    }

    \vfill 

    \subfloat[Average for short-CoT models]{
        \includegraphics[width=0.45\textwidth]{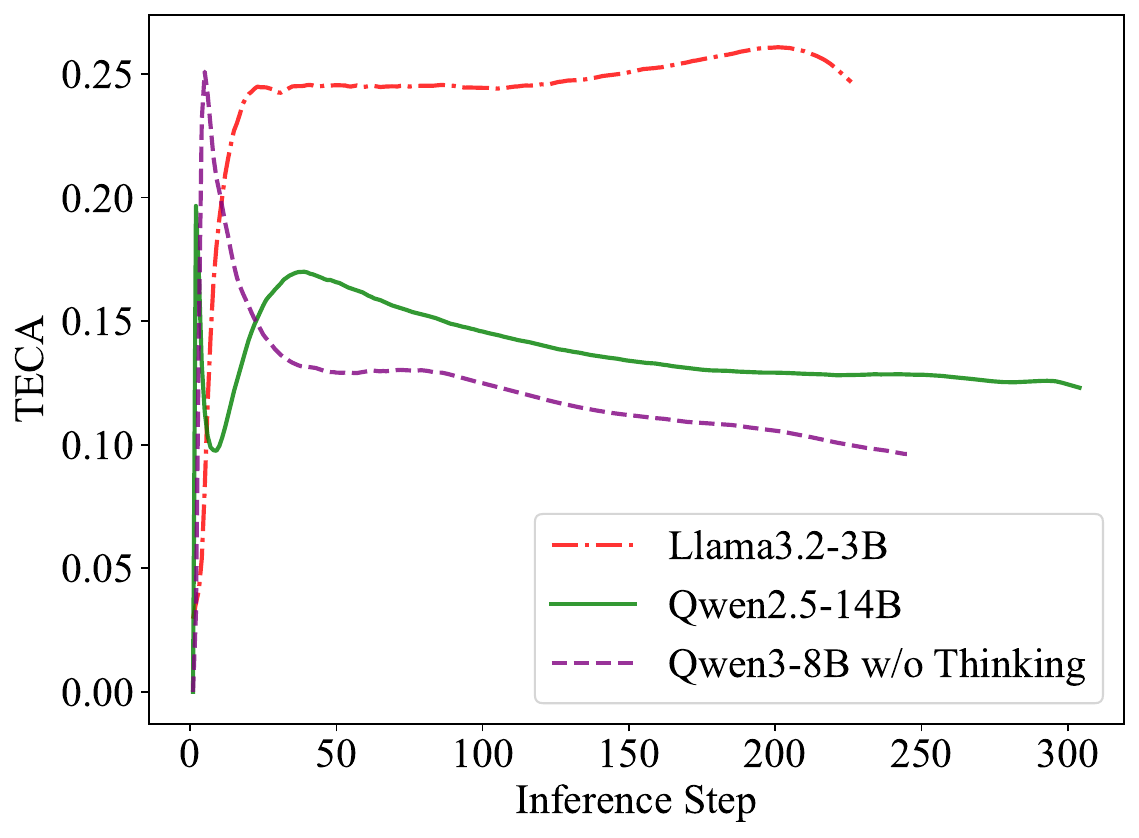}
        \label{fig1:c}
    }
    \hfil
    \subfloat[Average for a long-CoT model]{
        \includegraphics[width=0.45\textwidth]{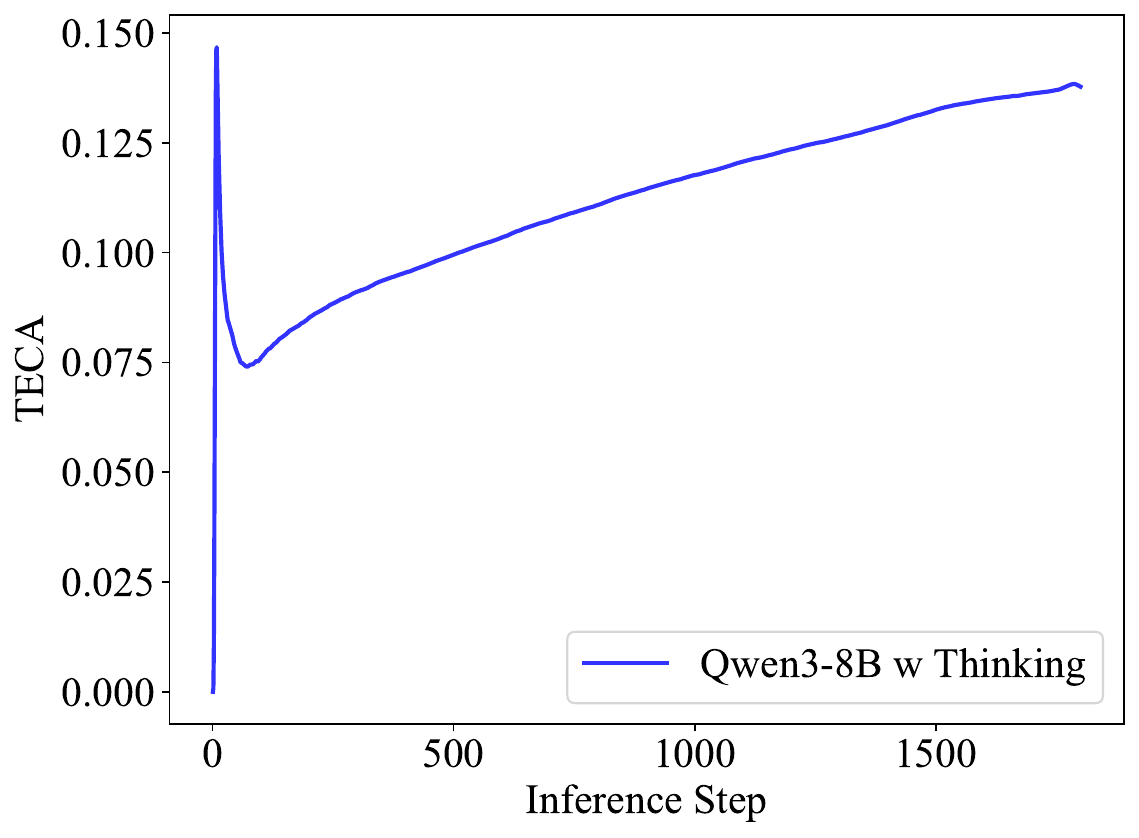}
        \label{fig1:d}
    }

    \caption{Token Entropy Cumulative Average curves in inference for four compared methods: Llama3.2-3B, Qwen2.5-14B and Qwen3-8B (without and with thinking versions). (a) and (b) correspond to one testing sample and (c) and (d) show the average results of 1000 samples, all from GSM8K dataset. The yellow star marks the step where the correct answer first appears.}
    \label{fig1:ATE_study}
\end{figure*}
 
\section{Preliminaries}
\subsection{Group Relative Policy Optimization (GRPO)}
The GRPO objective extends PPO by optimizing relative advantages within a group of responses and adding a KL penalty to a reference policy, making training more stable and robust to noisy absolute rewards as showed in Equation \ref{eq:grpo}:

\begin{equation}
\begin{aligned}
     \mathcal{J}_{\mathrm{GRPO}}(\theta)  = \mathbb{E} \bigl[ q \sim P(Q), \{o_i\}_{i=1}^G \sim \pi_{\theta_{\text{old}}}(O \mid q) \bigr] \\
     \Biggl\{ \frac{1}{G} \sum_{i=1}^G \Biggl( \min \biggl( \frac{\pi_\theta(o_i \mid q)}{\pi_{\theta_{\text{old}}}(o_i \mid q)} A_i, \qquad \qquad \\
    \text{clip} \Bigl( \frac{\pi_\theta(o_i \mid q)}{\pi_{\theta_{\text{old}}}(o_i \mid q)}, 1-\epsilon, 1+\epsilon \Bigr) A_i \biggr) \\
    - \beta \mathbb{D}_{\mathrm{KL}} \bigl( \pi_\theta \| \pi_{\text{ref}} \bigr) \Biggr) \Biggr\}.
\end{aligned}
\label{eq:grpo}
\end{equation}
Specifically, for each question $q$, GRPO uses the old policy $\displaystyle \pi_{\theta_{old}}$ to sample a group of outputs $\{o_1, o_2, \dots, o_G \}$  and then optimizes the policy model $\pi_\theta$ by maximizing the objective. The formula of KL penalty is defined as:
\begin{equation}
\displaystyle
\mathbb{D}_{KL}\!\left(\pi_\theta \,\|\, \pi_{\mathrm{ref}}\right)
= \frac{\pi_{\mathrm{ref}}(o_i \mid q)}{\pi_\theta(o_i \mid q)}
- \log \frac{\pi_{\mathrm{ref}}(o_i \mid q)}{\pi_\theta(o_i \mid q)} - 1 ,
\label{eq:kl}
\end{equation}
where $\epsilon$ and $\beta$ are hyper-parameters, and $A_i$ is the advantage, computed using a group of rewards $\{r_1, r_2, . . . , r_G \}$ which are the outputs within each group respectively. The $A_i$ is from the Equation \ref{eq:adv_cal}:

\begin{equation}
\displaystyle
A_i = \frac{r_i - \mathrm{mean}(\{r_1, r_2, \cdots, r_G\})}{\mathrm{std}(\{r_1, r_2, \cdots, r_G\})}.
\label{eq:adv_cal}
\end{equation}

In our work, we use the GRPO algorithm as the base RL train algorithm, and then modify the reward calculation for concise thinking.

\subsection{Token Entropy}
The definition of token entropy will follow Wang et al. ~\cite{wang8020Rule2025}. Token entropy is calculated by token generation distribution, independent of specific token. The calculation of token entropy essentially is the information entropy of the event of outputting a token. Here, all possible outcomes of the event correspond to the vocabulary. The formula of token entropy is defined as:

\begin{equation}
\begin{aligned}
    H_t :=  -\sum_{j=1}^{V} p_{t,j} \log p_{t,j}, \quad \text{where } (p_{t,1}, \dots, p_{t,V}) = \mathbf{p}_t \\
    = \pi_\theta(\cdot \mid \mathbf{q}, \mathbf{o}_{<t}) = \mathrm{Softmax}\left(\frac{\mathbf{Z}_t}{T}\right).
\end{aligned}
\label{eq:te}
\end{equation}

Here, $\pi_{\boldsymbol{\theta}}$ represents the LLM parameterized by $\displaystyle \boldsymbol{\theta}$. $\boldsymbol{Z}_t$, $t$, $V$, $T$ is output logits, inference step(the $t$-th token model generating), vocabulary size, decoding temperature respectively. We choose softmaxed with temperature logits as token probability $p_{t, j}$ which means the probability of model generating the $t$-th inference step, $j$-th token in vocabulary. Then, $H_t$ is sum of all information entropy in vocabulary. 

As described in \emph{A Mathematical Theory of Communication} ~\cite{entropy}, information entropy can represent the average level of uncertainty. While token entropy formula is based on the information entropy formula, so token entropy represents the uncertainty of next token prediction. \textbf{The higher the token entropy, the greater the uncertainty of the model in predicting the next token and vice versa}. This is a significant property about token entropy and LLM generation.

\section{Explore Briefly, Then Decide: RL with Cumulative Entropy Regulation}
\label{CER_method}
In this part, we present our Cumulative Entropy Regulation (CER) method for standard GRPO reinforcement learning pipeline. To this end, we first define a new metric Token Entropy Cumulative Average (TECA) and analyze its correlation with overthinking during inference. Based on our hypothesis, we introduce the specific manner (solution) to implement CER with TECA in the RL framework for efficient reasoning. 

\subsection{Metric: Token Entropy Cumulative Average}
\label{step entropy property}
According to previous research~\cite{wang8020Rule2025} as well as our understanding, in language models, token entropy may describe the uncertainty of next token prediction. In the process of reasoning, it indicates the \textbf{exploration extent} at the specific generation step, suggesting the local certainty of the thinking process. However, to optimize the thinking length and meanwhile tackle the overthinking issue, it is required to regulate the whole thinking process instead of single generation step. Therefore, based on token entropy, we propose a new metric \textbf{T}oken \textbf{E}ntropy \textbf{C}umulative \textbf{A}verage (\textbf{TECA}) that may describe exploration extent across the whole thinking process. Specifically, we define the TECA as the average of all of the token entropy up until the current inference step, showed in Equation \ref{eq:teca definition}:
\begin{equation}
\displaystyle \mathrm{TECA}_{t} := \frac{\displaystyle \sum_{1}^{t} H_t}{t},
\label{eq:teca definition}
\end{equation}
where $t$ denotes a specific inference step, also the index of the token generated. $H_t$ denotes the corresponding token entropy referring Equation~\ref{eq:te}. The definition explains that at specific inference step $t$, $\text{TECA}_t$ measures the average exploration extent of the thinking process in the past, which indicates the global certainty of the model on the output so far. The higher TECA is, the less certain the model is in one reasoning process.

\subsection{discovery: TECA and overthinking}
\label{CER method discovery}
To investigate the correlation between TECA and the thinking process, we examined samples from the GSM8K dataset ~\cite{gsm8k_dataset}. In Figure.~\ref{fig1:ATE_study}, we present the TECA curves for four test settings: one Llama and two Qwen models with thinking enabled or not for Qwen3. We show both a detailed single-sample result and the average results across 1,000 samples for all four test settings, considering only the process with correct answers. For our analysis, we normalized all model responses based on the number of inference steps to align the TECA results across different samples. This was achieved by using first-order linear interpolation to standardize the inference step dimension.

The TECA curves reveal clear patterns during the inference process. At the beginning, all four test settings show a sharp increase or wide fluctuation in TECA, which signals the ``\textbf{Exploration}'' stage. We believe this stage is a necessary part of the model's effort to find possible paths leading to the correct answer. The specific shape of the Exploration stage varies across models, which may be related to their unique training methods or model architectures. After this initial stage, the TECA for three of the short-thinking settings, \textit{Llama3.2-3B} ~\cite{llama3technicalreport}, \textit{Qwen3-8B without Thinking}, and \textit{Qwen2.5-14B} ~\cite{qwen2technicalreport, qwen2.5}, drops and remains flat (as shown in sub-figures (a) and (c)). This indicates that they have moved into the ``\textbf{Determination}'' stage, as a low TECA suggests the model is no longer generating new thinking paths and is instead focused on confirming the final answer.

However, a different pattern emerges from the \textit{Qwen3-8B} model with thinking mode enabled. Both the single-case and average results show the TECA continuing to rise, suggesting the model keeps exploring without ever entering a determination stage. As illustrated in sub-figure (b), the correct answer often appears very early at roughly one-third of the way through the inference process, yet the model continues to generate a long, redundant response. This continuous exploration results in an average response length (1897.36 tokens) that is far greater than that of the other models, including \textit{Llama3-8B} (242.113 tokens), \textit{Qwen2.5-14B} (312.153 tokens), and even \textit{Qwen3-8B} without its thinking mode (253.945 tokens). This comparison highlights how an unregulated reasoning process leads to overthinking. Most importantly, our analysis demonstrates a strong correlation between the TECA metric and the model's thinking stages. Therefore, we propose the hypothesis that over-exploration leads to the overthinking.

\subsection{proof: prove the hypothesis from two aspects}
\label{claimed hypothesis}
As discussed in Section \ref{step entropy property}, higher TECA indicates the uncertainty of the model in the whole reasoning process. 
Figure.~\ref{fig1:ATE_study} illustrates the TECA of \textit{Qwen3-8B} keeping rising during answering, whereas that of \textit{Llama3.2-3B} and \textit{Qwen2.5-14B} going up first, and remaining unchanged or decreasing instead after.
Previous study~\cite{wang8020Rule2025} discussed the existence of forking tokens during the reasoning process and pointed out the association between forking tokens with token entropy. Specifically, forking tokens usually correspond to high token entropy. Therefore, we can conclude that our TECA metric may increase when forking tokens appear. On one hand, the presence of forking tokens represents the beginning of model exploration; on the other hand, the continuous increase in TECA indicates the model's uncertainty in its response. Combining these two aspects, we can interpret the model's TECA curve as follows: as reasoning progresses, an increase in TECA indicates that the model is exploring answers, and the response at this stage is filled with uncertainty. A decrease in TECA, however, signifies that the model is becoming more confident in its current response and can output the final, definitive answer.

\begin{figure}
    \centering
    \includegraphics[width=1\linewidth]{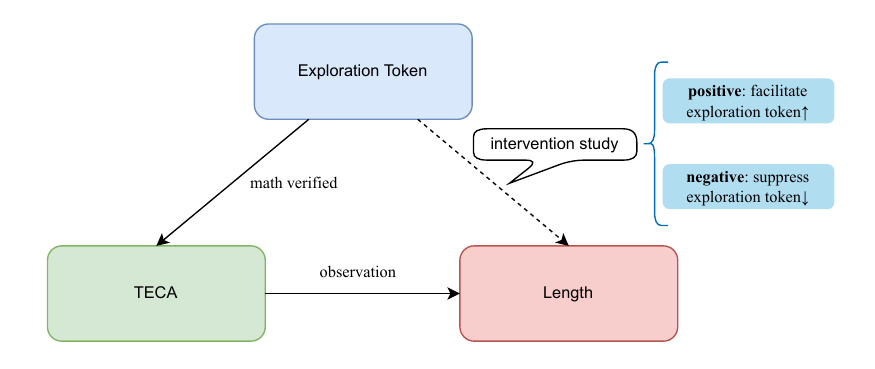}
    \caption{Causal diagnostic framework illustrating the relationships among exploration tokens, TECA, and response length.}
    \label{fig:Causal Diagnostic}
\end{figure}

From the perspective of causal diagnostic, we additionally examine the relationship between TECA trajectories and response length.
Figure \ref{fig:Causal Diagnostic} illustrates the overall framework. Our starting point is the observation that TECA is closely correlated with response length, and that TECA is mathematically derived from token entropy and the occurrence of exploration tokens. From TECA’s definition, it is straightforward to verify that the presence of high-entropy exploration tokens contributes directly to increases in TECA due to its cumulative-average formulation. Thus, the remaining key question is establishing the causal link between exploration tokens and response length. Two prior studies provide complementary intervention-based evidence: 

\begin{itemize}[topsep=0pt, partopsep=0pt, itemsep=2pt, parsep=2pt]
    \item \textbf{Positive intervention}: Wang et al. ~\cite{wang8020Rule2025} showed that optimizing the RLVR training procedure increases the number of exploration tokens, which in turn leads to longer responses.
    \item \textbf{Negative intervention}: Wang et al. ~\cite{wangWaitWeDont2025} demonstrated that suppressing rethinking/exploration tokens at the logit level results in shorter responses.
\end{itemize}

These findings offer convergent causal support for the claim that exploration tokens influence response length. Consequently, they also reinforce the causal relationship between TECA which is constructed from token entropy and response length. 

\subsection{Solution: Cumulative Entropy Regulation on exploration}
Based on the above arguments and proof, to mitigate the overthinking problem, we propose to regulate the model's exploration during training, but only suppress the model's excessive exploration, not the model's inherent exploratory ability. We expect the model to explore solutions for the problem and then gradually become certain about the final answer after finding the correct path. This process corresponds to the TECA curve first rising and then falling. To achieve this, we designed a TECA reward and a segmented reward mechanism based on the GRPO. TECA reward is used to suppress the over-exploration and segmented reward mechanism is used to reserve the model's exploratory ability.

\subsubsection{Reward Function}
We propose a combined reward function with two components: accuracy and TECA, aiming to 
suppress the model's excessive exploration and avoid to force the model to output only low-entropy tokens while sacrifice accuracy. 
We define our TECA reward as follows:
\begin{equation}
\displaystyle
r_{\mathrm{te}} = e^{-\mathrm{TECA}_{-1}} + 1,
\label{step entropy reward}
\end{equation}
where $\mathrm{TECA}_{-1}$ denotes the TECA value in last inference step during answering. 
By using $e^{-x}$, the TECA reward is constrained between 0 and 1 aligning with the accuracy reward while maintaining the inverse relationship of TECA with the reasoning efficiency, i.e., thinking length. We apply a trick in the reward function by adding 1 to the TECA reward to encourage the model lean to this part during optimization when multiple reward components are considered. 
The accuracy reward is defined as:
\begin{equation}
\displaystyle
r_{\mathrm{acc}} = 
\begin{cases}
0 & \text{if} ~~~ y \neq y_{\text{gt}} \\
1 & \text{if} ~~~ y = y_{\text{gt}}
\end{cases},
\label{accuracy reward function}
\end{equation}
where $y_\text{gt}$ and $y$ denote the ground truth and model prediction respectively.

\subsubsection{Segmented Reward Mechanism(SRM)}
An effective reward should suppress only the model's excessive exploration while ensure its exploratory ability remained. Specifically, the TECA reward is only activated when the model answers correctly, rather than being given regardless of correctness. This ensures that the model receives additional reward only when it successfully answers after sufficient exploration; otherwise, it should consider the accuracy reward only.
To this end, we design an effective segmented reward mechanism to delicately combine the two reward components, i.e., accuracy and TECA, as follows:
\begin{equation}
\displaystyle
r = 
\begin{cases}
    r_{\mathrm{acc}} & \text{if} ~~~ y \neq y_{\text{gt}}\\[6pt]
    \dfrac{r_{\mathrm{acc}} + r_{\mathrm{te}}}{2} & \text{if} ~~~ y = y_{\text{gt}}.
\end{cases}
\label{SRM function}
\end{equation}
After the model obtains the correct answer, a lower $\mathrm{TECA}_{-1}$ results in more rewards, thereby encouraging the model to learn to confirm its answer during the ``Determination Stage,'' manifested as a downward trend at the tail of the TECA curve instead of the original trend. We simply average the contribution of the two reward components, i.e., accuracy and TECA, to achieve the combined reward. Note that more sophisticated design could be considered here to integrate multiple components, we leave it for future study. To summarize, with the proposed segmented reward mechanism, reasoning models can learn to autonomously choose the extent of exploration to ensure both answering correctly and obtaining a higher TECA reward.

\begin{figure*}[t!] 
    \centering
    
    \subfloat[Qwen3-4B]{
        \begin{minipage}{\textwidth}
            \centering
            \includegraphics[width=0.31\textwidth]{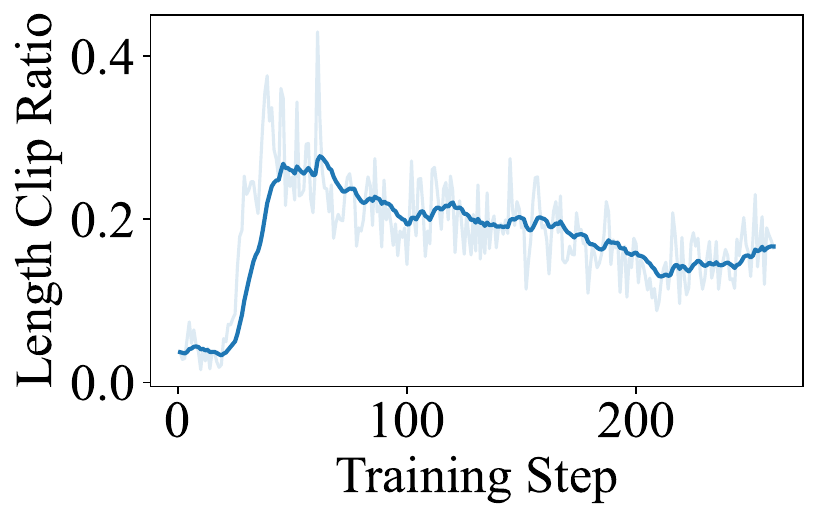}\hfill
            \includegraphics[width=0.31\textwidth]{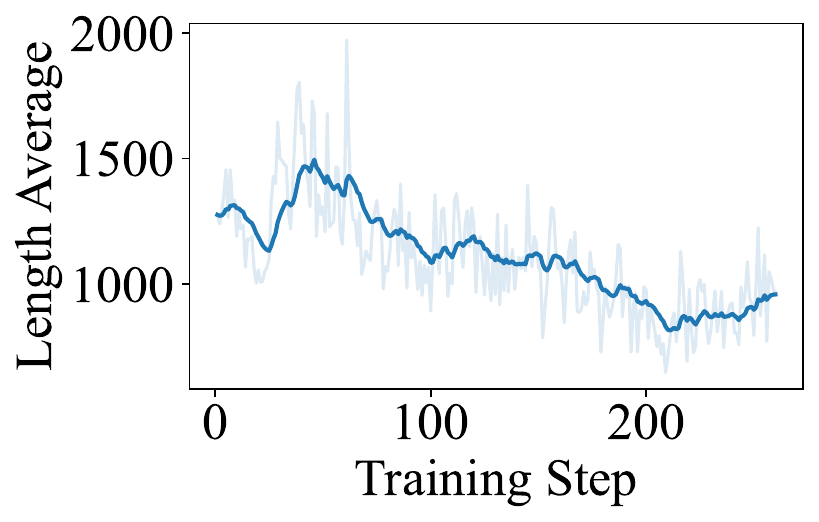}\hfill
            \includegraphics[width=0.31\textwidth]{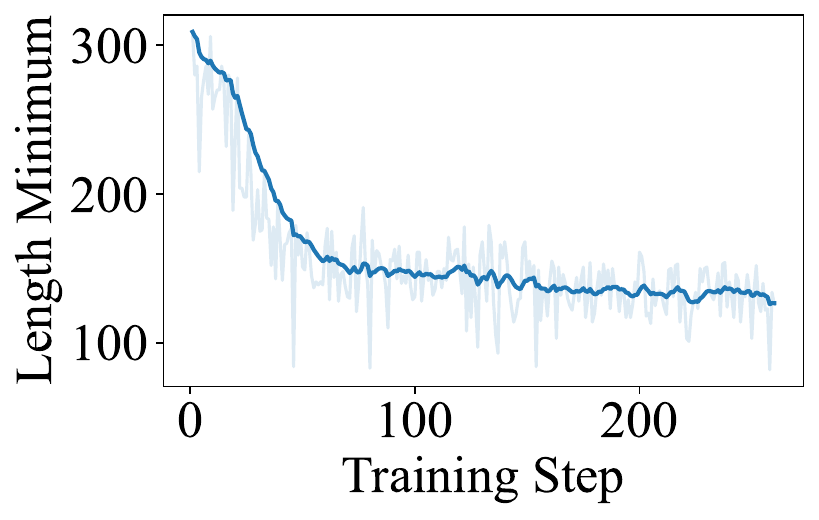}
        \end{minipage}
        \label{fig3:qwen3_4b}
    }


    \subfloat[Qwen3-8B]{
        \begin{minipage}{\textwidth}
            \centering
            \includegraphics[width=0.31\textwidth]{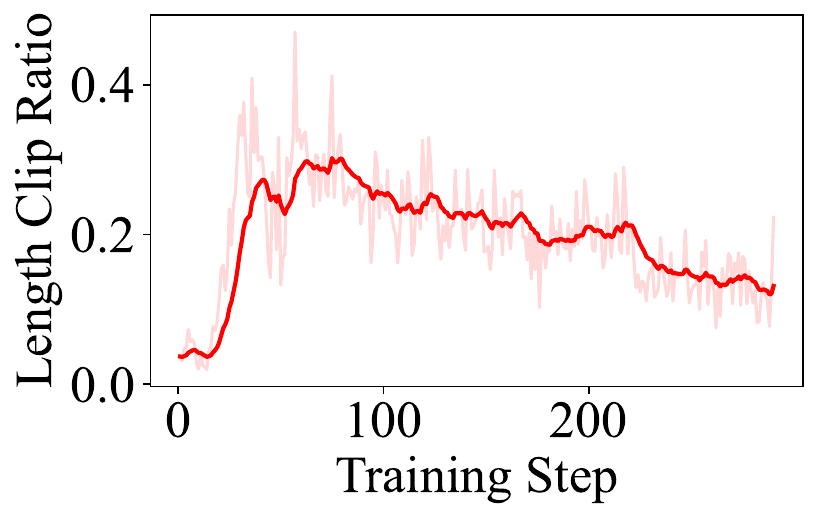}\hfill
            \includegraphics[width=0.31\textwidth]{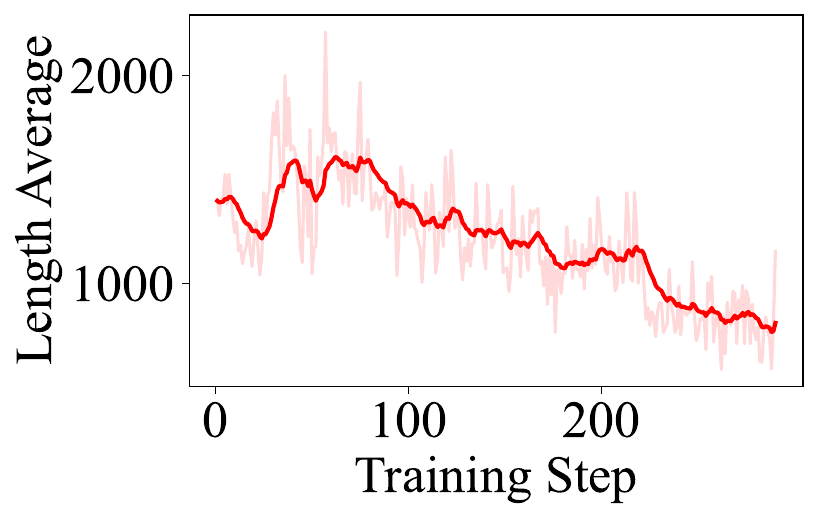}\hfill
            \includegraphics[width=0.31\textwidth]{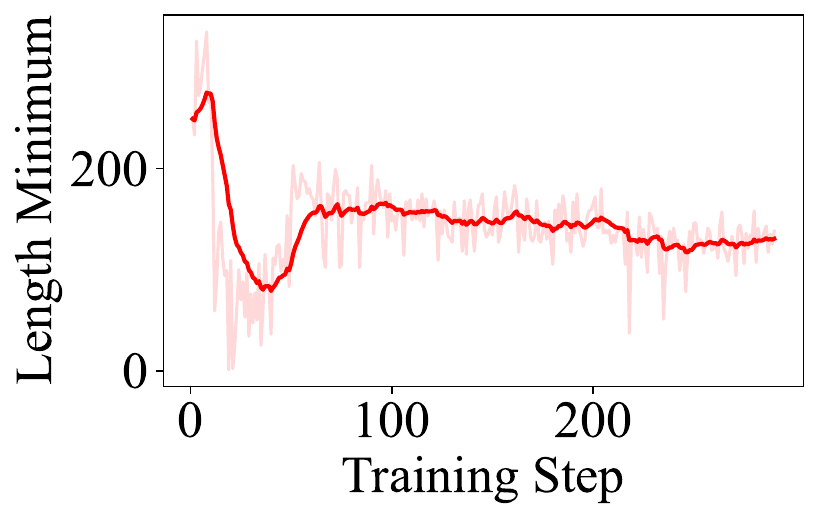}
        \end{minipage}
        \label{fig3:qwen3_8b}
    }
    
    \caption{Inference token length curve through CER training on two LLMs. 
    Left: Length Clip Ratio denoting the proportion of responses exceeding the max length of the model; 
    Middle: Average Length of all responses in a group; 
    Right: Minimum Length of all responses in a group.}
    \label{fig3:training process}
\end{figure*}

\section{Experiments}
We conduct extensive experiments on two LRMs for the math benchmarks to evaluate the effectiveness of our proposed reward function and learning pipeline. Specifically, we fine-tune Qwen3-4B and Qwen3-8B~\cite{qwen3technicalreport} using LoRA~\cite{hu2021lora} by GRPO~\cite{deepseekr1} with the VERL framework~\cite{sheng2024verlframework}. We used the GSM8K~\cite{gsm8k_dataset} training set (7473 samples) as the training data and trained them for 5 epochs.

\begin{figure}
    \centering
    \includegraphics[width=1\linewidth]{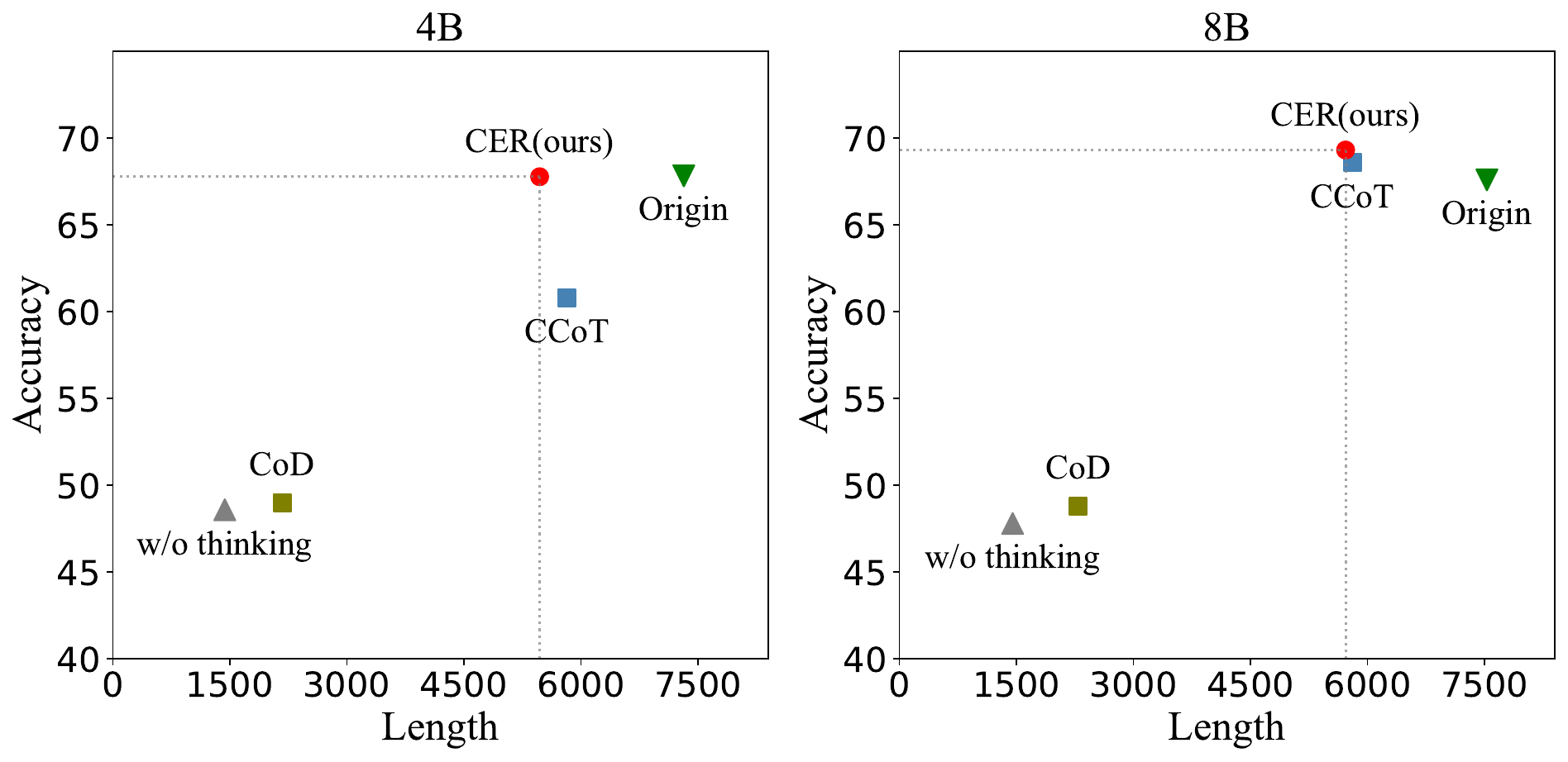}
    \caption{Performance comparison of different thinking methods on the Qwen3-4B and Qwen3-8B backbone models, measured by Accuracy (effectiveness, higher is better) and Token Length (efficiency, lower is better).}
    \label{fig:experiment scatter plot}
\end{figure}

Figure \ref{fig3:training process} illustrates the development of the generated token length in inference within the group during the training process. From the figure we can observe continuous dropping patterns for two observed metrics, i.e., Length Clip Ratio and Average Length, showing our learning approach can effectively decrease the token length for average and particularly long thinking process cases. Meanwhile, we can also observe that Minimum Length has a clear lower bound as the curve stays flat along with the training process, suggesting the thinking process will not be unlimited compressed. 
Such training patterns demonstrate the effectiveness of our CER in terms of suppressing excessive exploration while preserving necessary exploration ability of LLMs in reasoning, thereby tackling the overthinking issue. The quantitative results on math benchmarks with two models (Qwen3-4B and Qwen3-8B) presented in Table~\ref{tab:CER_results} demonstrate the effectiveness and adaptability of our proposed CER method. First, the data shows that CER successfully transforms models prone to overthinking into more efficient reasoning models. Compared to the backbone model, CER significantly reduces the average response length across all four tested datasets, simplifying the thought process without compromising performance. Second, when compared to other existing methods like CoD and CCoT, CER achieves superior accuracy while also reducing response length more effectively. This highlights CER's ability to train models that are not only more efficient but also more accurate, demonstrating its overall effectiveness.

\begin{figure*}[t!] 
    \centering
    
    \subfloat[Qwen3-4B]{
        \begin{minipage}{0.85\textwidth} 
            \centering
            \includegraphics[width=0.45\textwidth]{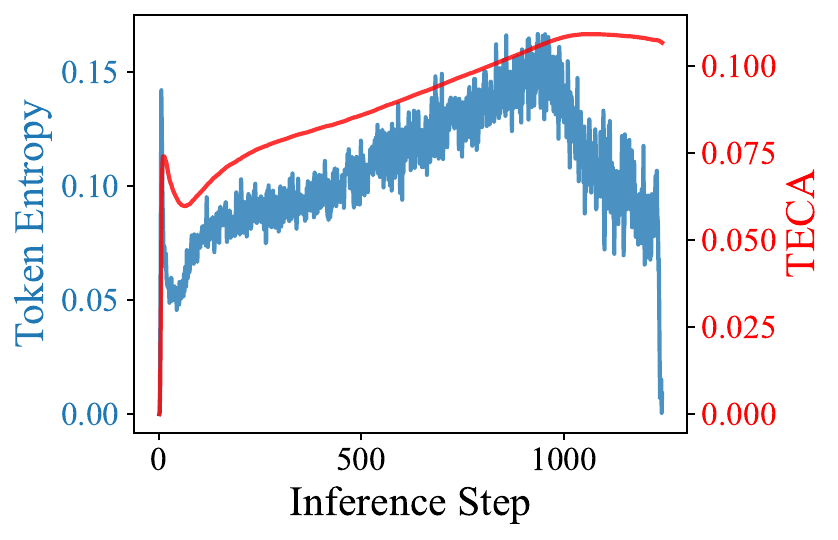}
            \hfil 
            \includegraphics[width=0.45\textwidth]{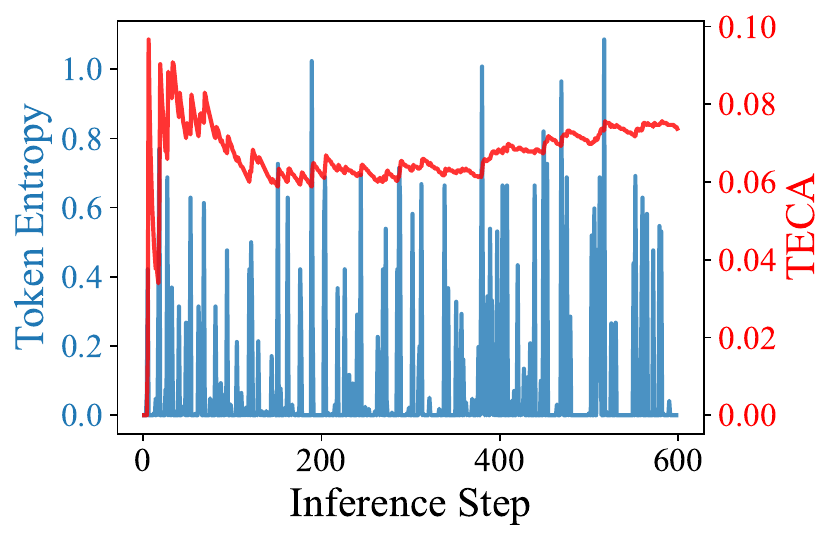}
        \end{minipage}
        \label{fig4:qwen3_4b_origin}
    }


    \subfloat[Qwen3-4B-CER (ours)]{
        \begin{minipage}{0.85\textwidth}
            \centering
            \includegraphics[width=0.45\textwidth]{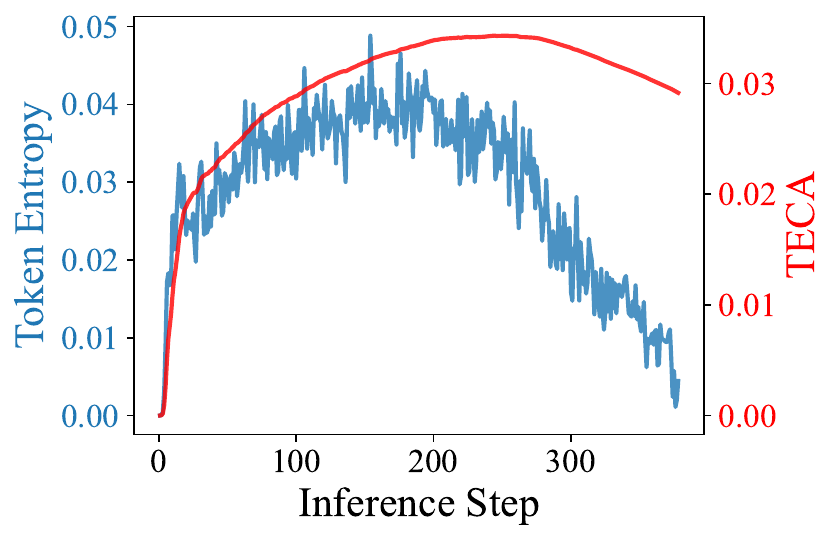}
            \hfil
            \includegraphics[width=0.45\textwidth]{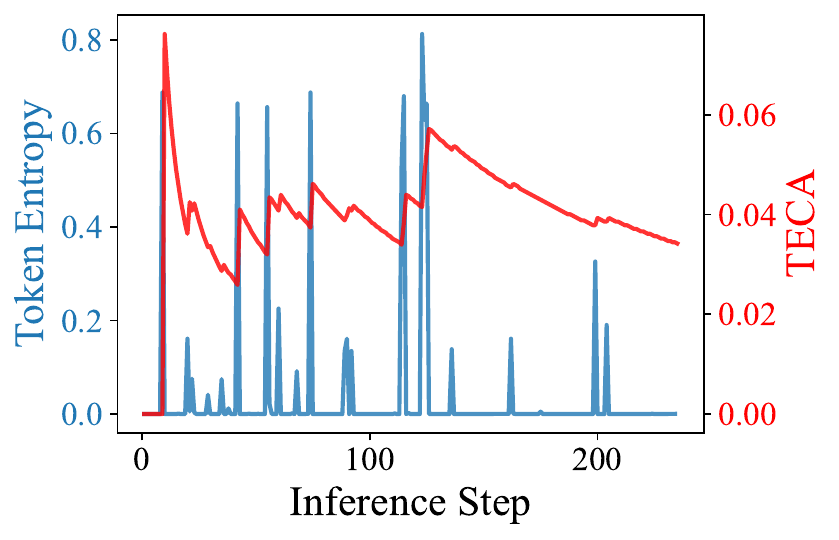}
        \end{minipage}
        \label{fig4:qwen3_4b_trained}
    }
    
    \caption{Token Entropy and TECA curves in inference for the reasoning models without and with our CER training. Left: average results for 1000 samples; Right: one case results.}
    \label{fig4:MECA_Comparison}
\end{figure*}

\begin{table*}[ht]
\centering
\caption{Performance of different methods on various math benchmarks. $\Delta$len indicates the percentage reduction in the length of the generated answer compared to the origin method (w thinking).}
\label{tab:CER_results}
\resizebox{\textwidth}{!}{
\begin{tabular}{lcccccccccccccc}
\toprule
\multirow{2}{*}{\textbf{Method}} & \multicolumn{3}{c}{\textbf{GSM8K}} & \multicolumn{3}{c}{\textbf{MATH500}} & \multicolumn{3}{c}{\textbf{AIME24}} & \multicolumn{3}{c}{\textbf{AIME25}} & \multicolumn{2}{c}{\textbf{Average}} \\
\cmidrule(lr){2-4} \cmidrule(lr){5-7} \cmidrule(lr){8-10} \cmidrule(lr){11-13} \cmidrule(lr){14-15}
& ACC & LEN & $\Delta$LEN & ACC & LEN & $\Delta$LEN & ACC & LEN & $\Delta$LEN & ACC & LEN & $\Delta$LEN & ACC & LEN \\
\midrule
\rowcolor{lightbg}
\multicolumn{15}{c}{\textbf{Qwen3-4B}} \\
\midrule
w thinking & 92.80 & 1348.59 & - & 65.20 & 4458.60 & - & 64.44 & 11343.57 & - & 48.89 & 12119.62 & - & 67.83 & 7317.59 \\
w/o thinking & 86.50 & 260.96 & 80.65\% & 61.20 & 846.35 & 81.02\% & 26.67 & 2132.2 & 78.34\% & 20.00 & 2503.93 & 82.84\% & 48.59 & 1435.86 \\

CoD & 93.30 & 385.50 & 71.41\% & 52.60 & 1159.73 & 73.99\% & 23.30 & 3607.83 & 68.19\% & 26.70 & 3535.23 & 70.83\% & 48.98 & 2172.07 \\
CCoT & 82.56 & 616.42 & 54.29\% & 64.00 & 2401.94 & 46.13\% & 56.67 & 9491.87 & 16.32\% & 40.00 & 10775.93 & 11.09\% & 60.81 & 5821.54 \\
\textbf{CER (ours)} & 94.09 & 391.08 & 71.00\% & 64.80 & 2708.65 & 39.25\% & 61.11 & 9215.77 & 18.76\% & 51.11 & 9565.64 & 21.07\% & 67.78 & 5470.29 \\
\midrule
\rowcolor{lightbg}
\multicolumn{15}{c}{\textbf{Qwen3-8B}} \\
\midrule
w thinking & 94.62 & 1491.38 & - & 65.80 & 4669.74 & - & 63.33 & 11247.68 & - & 46.67 & 12708.16 & - & 67.60 & 7529.24 \\
w/o thinking & 88.86 & 272.02 & 79.83\% & 59.00 & 837.16 & 81.22\% & 20.00 & 2399.13 & 80.10\% & 23.33 & 2300.13 & 80.69\% & 47.80 & 1452.11 \\
CoD & 94.40 & 415.80 & 72.12\% & 60.80 & 1391.22 & 70.21\% & 20.00 & 3657.57 & 67.48\% & 20.00 & 3709.20 & 70.81\% & 48.80 & 2293.45 \\
CCoT & 92.49 & 739.05 & 50.45\% & 65.20 & 2761.19 & 40.87\% & 63.33 & 9286.63 & 17.44\% & 53.33 & 10438.07 & 17.86\% & 68.59 & 5806.23 \\
\textbf{CER (ours)} & 92.57 & 668.06 & 55.21\% & 65.80 & 3140.04 & 32.76\% & 65.56 & 9171.56 & 18.46\% & 53.33 & 9894.51 & 22.14\% & 69.32 & 5718.54 \\
\bottomrule
\end{tabular}
}
\label{benchmark result}
\end{table*}

A particularly important finding is the adaptivity of CER. Our method achieves a much larger reduction in response length on simpler datasets like GSM8K than on more complex ones, such as the two AIME datasets. This crucial observation confirms that CER does not simply shorten the thought process indiscriminately; instead, it adaptively adjusts the length of the reasoning chain based on the complexity of the problem. This precisely addresses the core issue of ``overthinking'' by only suppressing unnecessary reasoning steps, validating our central claim that CER promotes efficient, targeted thinking rather than just abbreviated thinking.
\section{Discussion}
\subsection{Thinking Process Analysis}
In Section~\ref{claimed hypothesis}, we have explained the correlation between TECA and overthinking, along with the logic for using TECA as an indicator of over-exploration. A primary goal of our CER is to train the model to stabilize its answers after sufficient, but not redundant, exploration. This is the core principle behind our ``Explore Briefly, Then Decide'' paradigm. To explore whether CER effectively constrains the thinking process and mitigates overthinking, we analyzed token entropy and TECA values across inference steps. The average results from GSM8K test split(1319 samples) and a representative case from Qwen3-4B model are presented in Figure~\ref{fig4:MECA_Comparison}. 
First, by examining  the single case results from two models, we can observe that the Qwen3-4B-CER model exhibits fewer high token entropy peaks as the thinking process going. This suggests that the model is generating fewer ``forking'' tokens and is gradually converging toward a final determination. In contrast, the original Qwen3-4B model shows no clear change in token entropy or TECA during the final stages of the reasoning process, indicating a lack of convergence. 

Similar patterns are evident in the average results. The model trained with CER shows a clear increase in TECA at the beginning of the inference, indicating a deliberate and considerable exploration phase, suggesting our CER does not harm the critical exploration ability of the model. More importantly, we can observe over-exploration is effectively suppressed, as evidenced by a clear drop in TECA in the latter part of the inference. This demonstrates that the thinking process is reaching a determination stage and is actively avoiding overthinking. This behavior directly supports our claim that CER specifically targets and suppresses only the unnecessary ``overthinking'' portions of the reasoning process, rather than simply shortening the entire chain.

\subsection{Ablation Study}
For modular ablation, the CER framework consists of two key modules: (1) the TECA reward and (2) the SRM. Because the SRM is defined based on the TECA reward, the ablation mainly examines the effect of removing SRM and the effect of removing both TECA and SRM.

The results in Table \ref{modular ablation result} show that removing both TECA and SRM yields higher accuracy than full CER, but also leads to substantially longer responses. When only the SRM is removed, we observe a reduction in output length compared with the original model; however, accuracy on easy problems drops noticeably, while accuracy on hard problems shows a slight improvement. The corresponding change in length is also smaller than that achieved by full CER.

We consider the minor fluctuations on hard problems to be expected. This is related to the training strategy of CER: the model is encouraged to determine an answer and maintain confidence once it reaches a stable conclusion. This mechanism can occasionally lead the model to converge to an incorrect answer, which in turn introduces small variations in accuracy, particularly on more challenging tasks. Nonetheless, this effect is limited and does not alter the overall trends.
\begin{table}[htbp]
\centering
\caption{Modular ablation of Qwen3-4B}
\small 
\setlength{\tabcolsep}{3pt} 
\begin{tabular*}{\columnwidth}{@{\extracolsep{\fill}} lcccc}
\toprule
\multirow{2}{*}{\textbf{Module Variant}} & \multicolumn{2}{c}{\textbf{GSM8K}} & \multicolumn{2}{c}{\textbf{AIME24}} \\
\cmidrule(lr){2-3} \cmidrule(lr){4-5}
& \textbf{ACC} & \textbf{LEN} & \textbf{ACC} & \textbf{LEN} \\
\midrule
CER & 94.09 & 391.08 & 61.11 & 9215.77 \\
remove-SRM & 91.89 & 542.61 & 62.22 & 8750.14 \\
acc-only & 95.22 & 887.07 & 64.44 & 10953.92 \\
\bottomrule
\end{tabular*}
\label{modular ablation result}
\end{table}

For temperature ablation, adjusting the rollout temperature directly affects the degree of exploration during training: higher temperatures yield more stochastic outputs, which expand the exploration space and lead to correspondingly larger parameter update steps. As shown in Table \ref{temperature ablation result}, for easy problems, increasing the temperature generally enhances the length reduction effect, while also yielding modest improvements in accuracy.

For hard problems, the length reduction is less sensitive to temperature. This is expected, as solving these tasks inherently requires substantial reflection and exploration regardless of the sampling stochasticity. Interestingly, the accuracy at temperature = 0.8 is noticeably higher than at other temperatures. We attribute this to the fact that lower output randomness slows the model’s acquisition of the targeted reasoning pattern. At this stage, the model has not fully learned the “explore briefly, then decide” behavior, which results in higher accuracy but also significantly longer responses.

\begin{table}[htbp]
\centering
\caption{Temperature ablation of Qwen3-4B}
\small
\setlength{\tabcolsep}{3pt}
\begin{tabular*}{\columnwidth}{@{\extracolsep{\fill}} lcccc}
\toprule
\multirow{2}{*}{\textbf{Temperature}} & \multicolumn{2}{c}{\textbf{GSM8K}} & \multicolumn{2}{c}{\textbf{AIME24}} \\
\cmidrule(lr){2-3} \cmidrule(lr){4-5}
& \textbf{ACC} & \textbf{LEN} & \textbf{ACC} & \textbf{LEN} \\
\midrule
1.5 & 94.09 & 391.08 & 61.11 & 9215.77 \\
1.2 & 93.71 & 456.44 & 61.11 & 8703.79 \\
0.8 & 93.71 & 606.27 & 66.67 & 9646.43 \\
\bottomrule
\end{tabular*}
\label{temperature ablation result}
\end{table}

\subsection{CER indeed suppresses useless branches}
CER truly decreases the response length substantially, however whether CER suppresses useless branches or simply cuts off exploration globally should be verified. So, we conduct the following Pass@k study. Here, the length is aggregated across k samples averagely in the Pass@k. If CER truncated important exploration, the overall ability to find the correct answer across multiple samples (Pass@k) should drop significantly. Table \ref{pass@k study result} shows that Pass@k remains robust, despite the substantial length reduction. This demonstrates that CER successfully retains the model's core ability to find the correct reasoning path while eliminating unnecessary detours and post-convergence chatter.

\begin{table}[htbp]
\centering
\caption{Performance comparison of Qwen3-4B with different pass@k settings}
\small
\setlength{\tabcolsep}{0pt} 
\begin{tabular*}{\columnwidth}{@{\extracolsep{\fill}} lcccc}
\toprule
\multirow{2}{*}{\textbf{Method}} & \multicolumn{2}{c}{\textbf{GSM8K}} & \multicolumn{2}{c}{\textbf{AIME24}} \\
\cmidrule(lr){2-3} \cmidrule(lr){4-5}
& \textbf{ACC} & \textbf{LEN} & \textbf{ACC} & \textbf{LEN} \\
\midrule
origin-pass@4  & 96.29 & 1373.61 & 73.33 & 11199.52 \\
CER-pass@4     & 95.98 & 561.93  & 76.67 & 9325.61  \\
\midrule
origin-pass@8  & 96.89 & 1359.78 & 83.33 & 11328.77 \\
CER-pass@8     & 96.36 & 571.70  & 83.33 & 9272.23  \\
\midrule
origin-pass@32 & 97.50 & 1357.55 & 83.33 & 11270.63 \\
CER-pass@32    & 97.19 & 567.26  & 86.67 & 9158.83  \\
\midrule
origin-pass@64 & 97.80 & 1357.17 & 86.67 & 11294.85 \\
CER-pass@64    & 97.42 & 568.77  & 83.33 & 9039.03  \\
\bottomrule
\end{tabular*}
\label{pass@k study result}
\end{table}

\subsection{Evidence of exploration ability after trained}
\begin{figure}[t!]
    \centering
    \subfloat[Qwen3-4B]{
        \includegraphics[width=0.45\columnwidth]{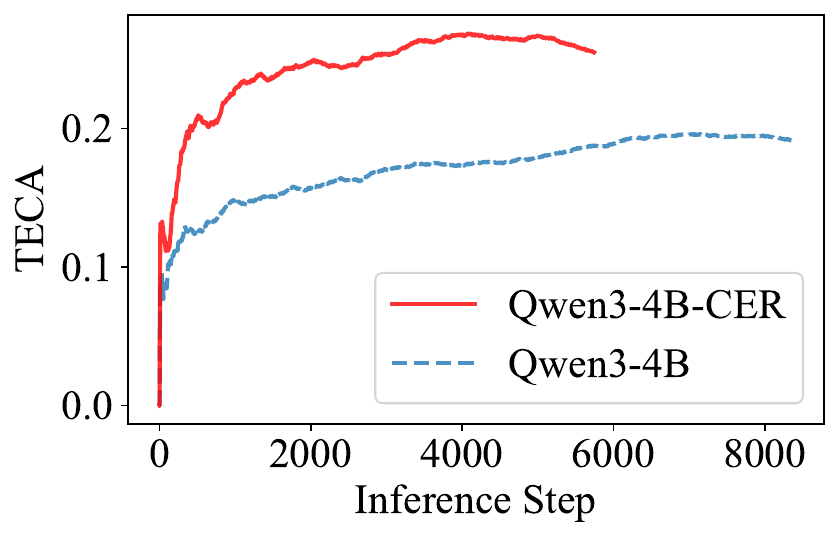}
        \label{apdx_fig2:a}
    }
    \hfil 
    \subfloat[Qwen3-8B]{
        \includegraphics[width=0.45\columnwidth]{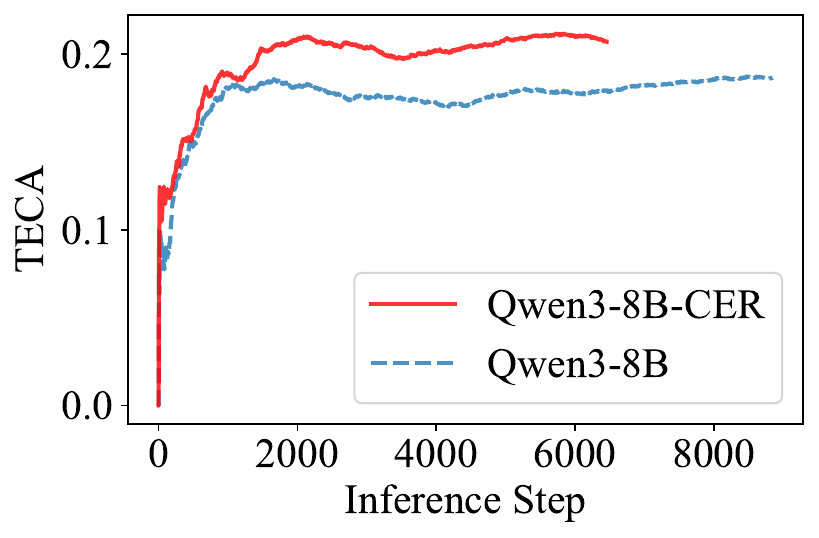}
        \label{apdx_fig2:b}
    }
    \caption{Comparison between origin and trained TECA for Qwen3-4B and Qwen3-8B in AIME dataset.}
    \label{apdx_fig2: Comparison between origin and trained TECA}
\end{figure}

For Llama3.2-3B and Qwen2.5-14B, there is a clear difference in the TECA trend between correctly and incorrectly answered questions. In the ``Determination Stage,'' if these models answer incorrectly, they show an obvious upward trend, meaning the model's uncertainty about the answer increases noticeably. For the overthinking model Qwen3-8B, regardless of whether the answer is correct or not, there is no significant difference in these trends; TECA continues to rise steadily, with incorrect answers showing a more pronounced and uncertain increase. This indicates that for non-overthinking models, there is a certain correlation between whether the model answers correctly and the rise of TECA, while for reasoning models, due to the existence of forking tokens, a rise in TECA is inevitable. However, the problem is that it does not decrease, so we aim to mimic non-overthinking models by ensuring that TECA decreases at the final step. Therefore, CER is designed to suppress TECA at the last step, enabling the model to learn this trend.

As showed in Figure \ref{apdx_fig2: Comparison between origin and trained TECA}, after CER, the model's inherent exploration ability is not impaired; on the contrary, it even exhibits a stronger desire for exploration than the original model. The trained model's TECA achieves higher value on the AIME dataset while being significantly shorter in length than the original model. This also demonstrates that the trained model spontaneously reduces a large amount of redundant reasoning and decreases inference depth. 

\section{Conclusion}
This paper addressed the overthinking problem in large language models by introducing a new metric, Token Entropy Cumulative Average (TECA), to measure a model's reasoning exploration. A novel paradigm, ``Explore Briefly, Then Decide'' was further proposed, which uses a Cumulative Entropy Regulation (CER) mechanism to guide models toward efficient thinking. Experiments on math benchmarks showed our method effectively mitigates overthinking. In future, we plan to explore more sophisticated reward functions and test our approach on a wider range of LLMs with different reasoning mechanisms to further validate its effectiveness.

\section*{Acknowledgments}
This work is supported by the Central Guidance on Local Science and Technology Development Fund of Shanghai City (YDZX20253100002004), National Key Research and Development Program of China (2025YFF0522500), 
the Fundamental and Interdisciplinary Disciplines Breakthrough Plan of the Ministry of Education of China (JYB2025XDXM116), and the Fundamental Research Funds for the Central Universities.

\bibliographystyle{IEEEtran}
\bibliography{IEEEabrv, TPAMI}


\begin{IEEEbiography}[{\includegraphics[width=1in,height=1.25in,clip,keepaspectratio]{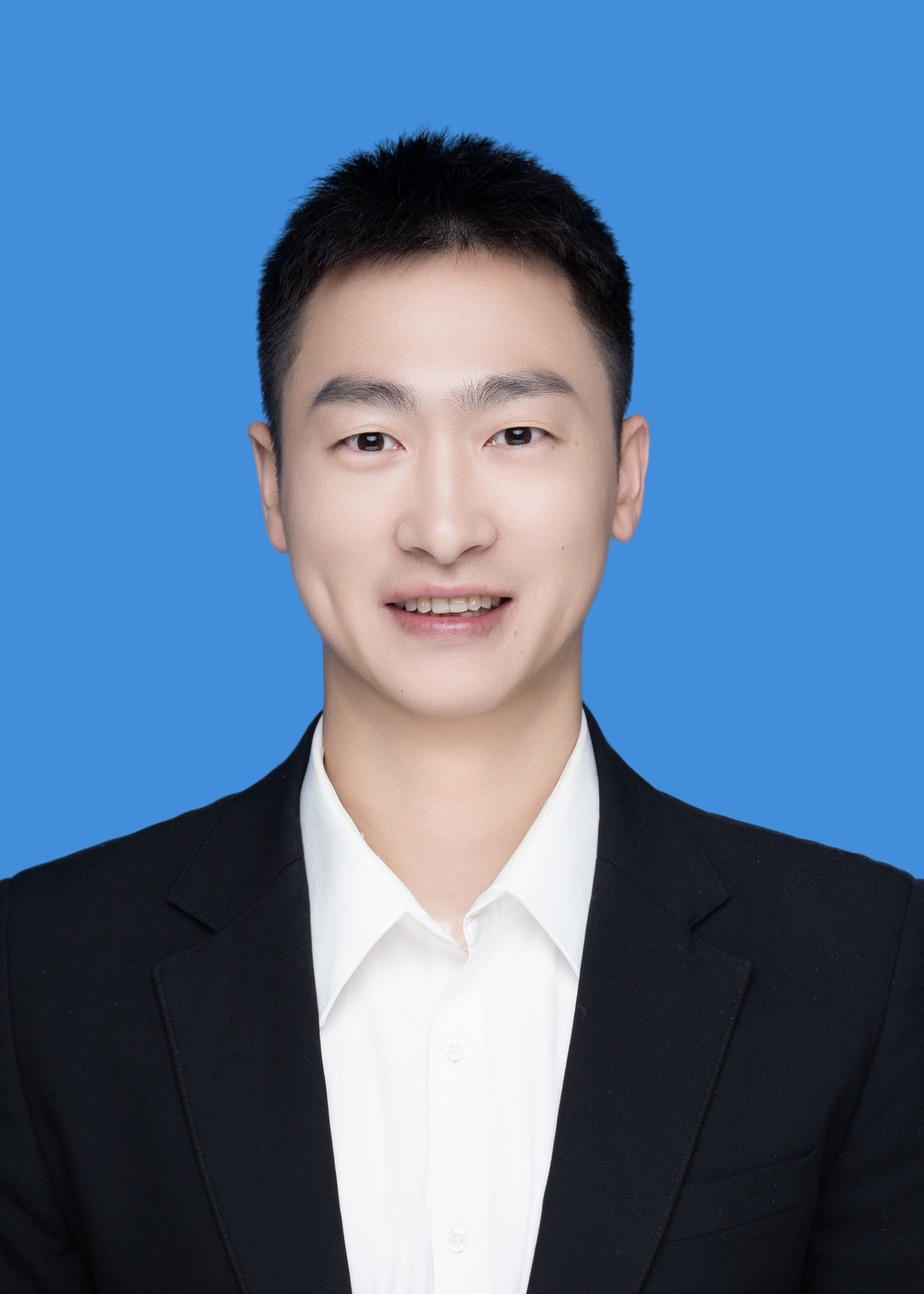}}]{Yi Bin} received the PhD degree from the University of Electronic Science and Technology of China (UESTC), in 2020. He is currently with Tongji University, Shanghai, China. His research interests include multimedia analysis, multimodal reasoning, and MLLMs.
\end{IEEEbiography}

\begin{IEEEbiography}[{\includegraphics[width=1in,height=1.25in,clip,keepaspectratio]{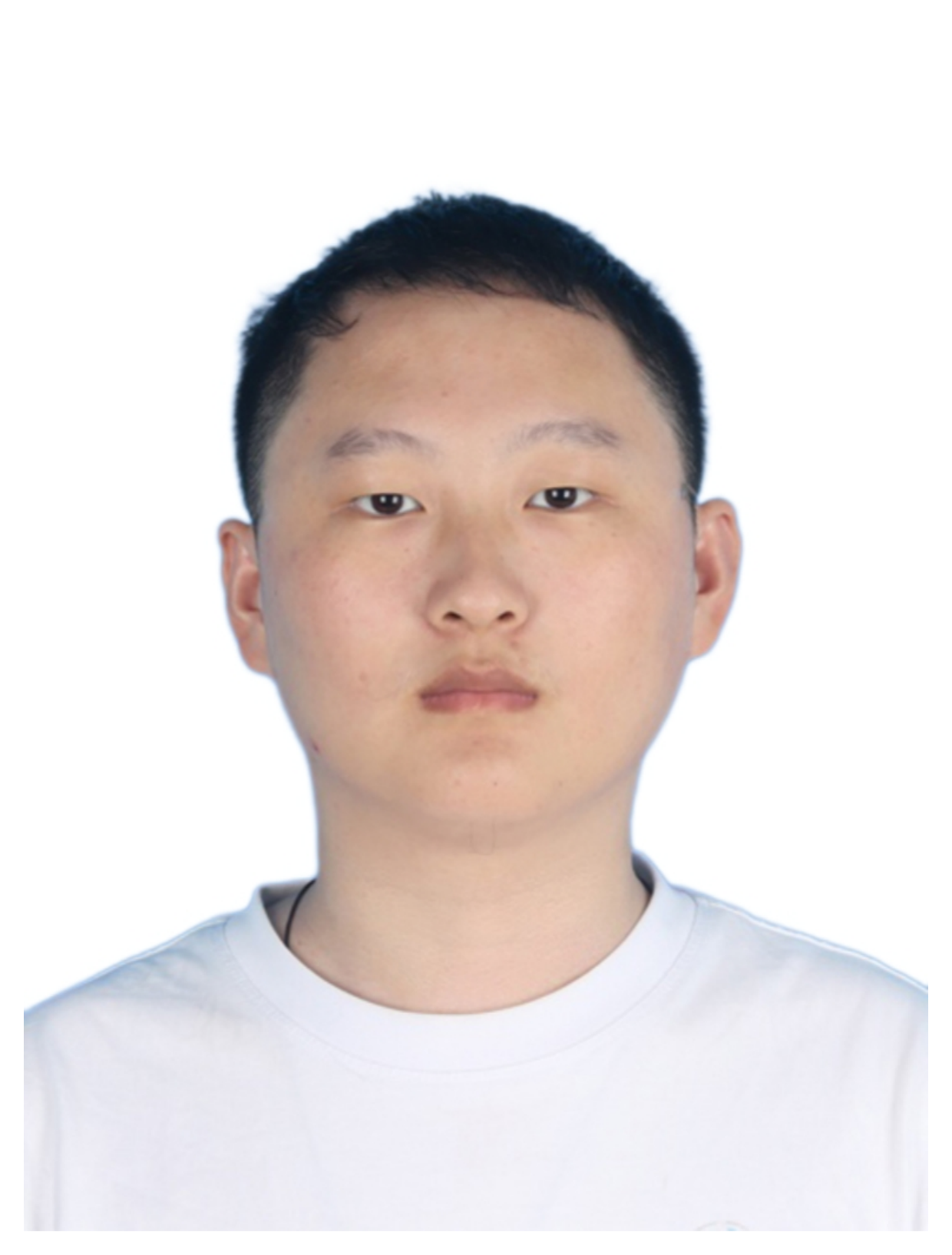}}]{Tianyi Jiang} received the BE degree in network engineering from Shanghai Maritime University, China,
in 2025. He is currently working towards the PhD degree in computer science with Tongji University, shanghai, China. His research interests include multimodal LLM, natural language processing, spatial intelligence.
\end{IEEEbiography}

\begin{IEEEbiography}[{\includegraphics[width=1in,height=1.25in,clip,keepaspectratio]{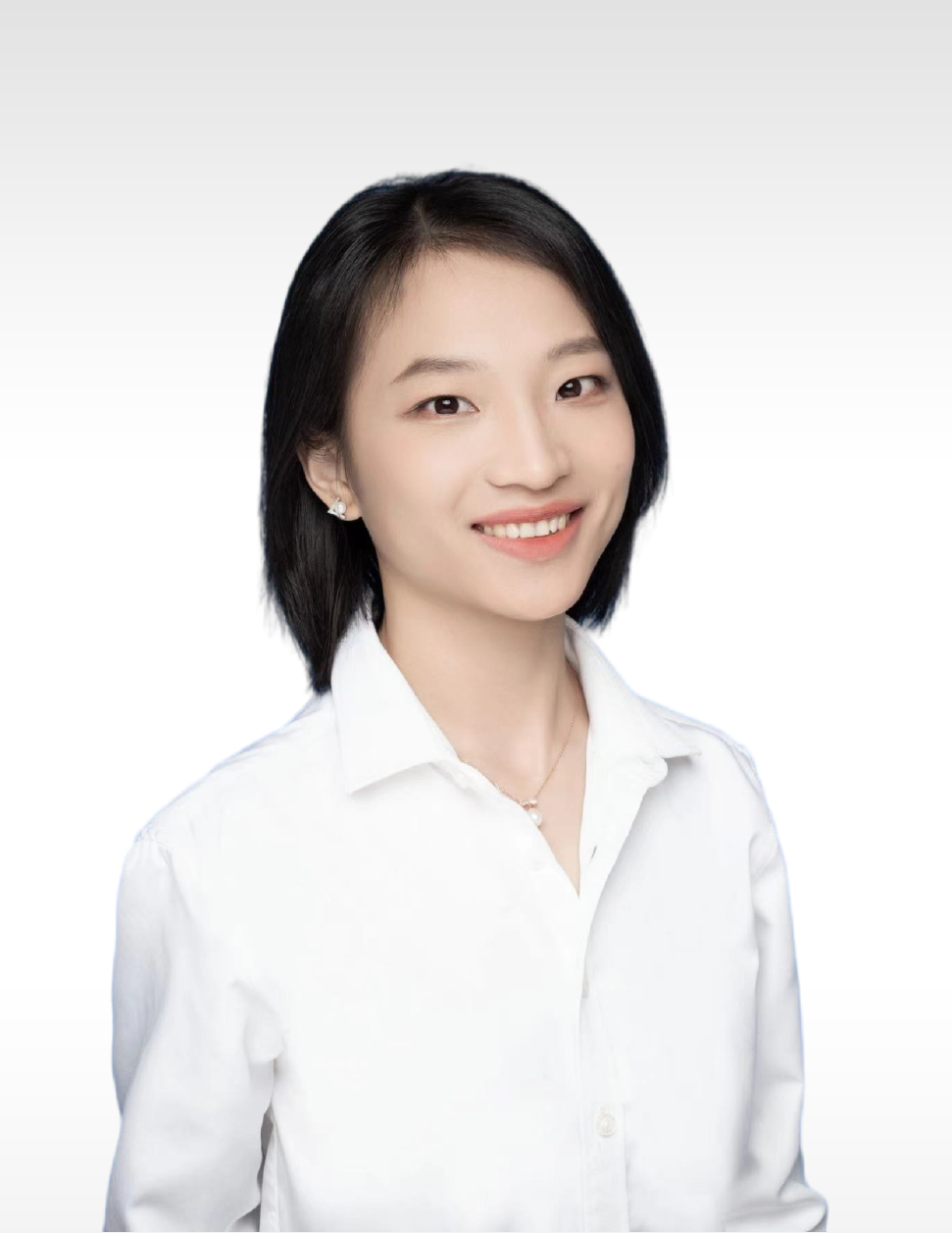}}]{Yujuan Ding} received the PhD degree from the Hong Kong Polytechnic University (PolyU). She is currently with PolyU as a research assistant professor. Her research interests include Retrieval-Augmented Generation (RAG), Multimedia Analysis, and Recommender Systems.
\end{IEEEbiography}

\begin{IEEEbiography}[{\includegraphics[width=1in,height=1.25in,clip,keepaspectratio]{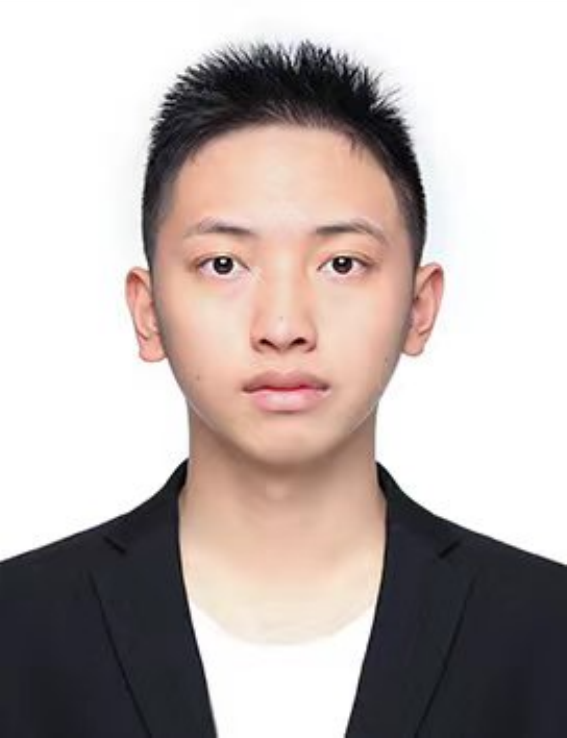}}]{Kainian Zhu} is currently working toward the B.S. degree in computer science and technology at Shanghai University of Electric Power, Shanghai, China. Since December 2024, he has been a Research Intern with Tongji University, Shanghai, China. He was a recipient of the Silver Medal in the International Collegiate Programming Contest (ICPC) Asia-East Continent Regional Contest. His research interests include algorithm design, reinforcement learning, and intelligent agents.
\end{IEEEbiography}

\begin{IEEEbiography}[{\includegraphics[width=1in,height=1.25in,clip,keepaspectratio]{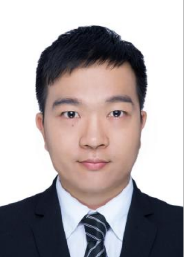}}]{Fei Ma} is currently a researcher at Guangdong Laboratory of Artificial Intelligence and Digital Economy (SZ). Before that, he received the B.S. degree in Communication Engineering from University of Electronic Science and Technology of China in 2017 and the Ph.D. degree in Information and Communication Engineering from Tsinghua University in 2022. So far, he has published over 40 papers in top-tier journals such as IEEE TPAMI, IEEE TMC, IEEE TAFFC, and IEEE TIE, as well as in prestigious conferences including NeurIPS, ICLR, ACL, AAAI, IJCAI, WWW, and ACM MM. His research interests include generative AI and multimodal learning.
\end{IEEEbiography}

\begin{IEEEbiography}[{\includegraphics[width=1in,height=1.25in,clip,keepaspectratio]{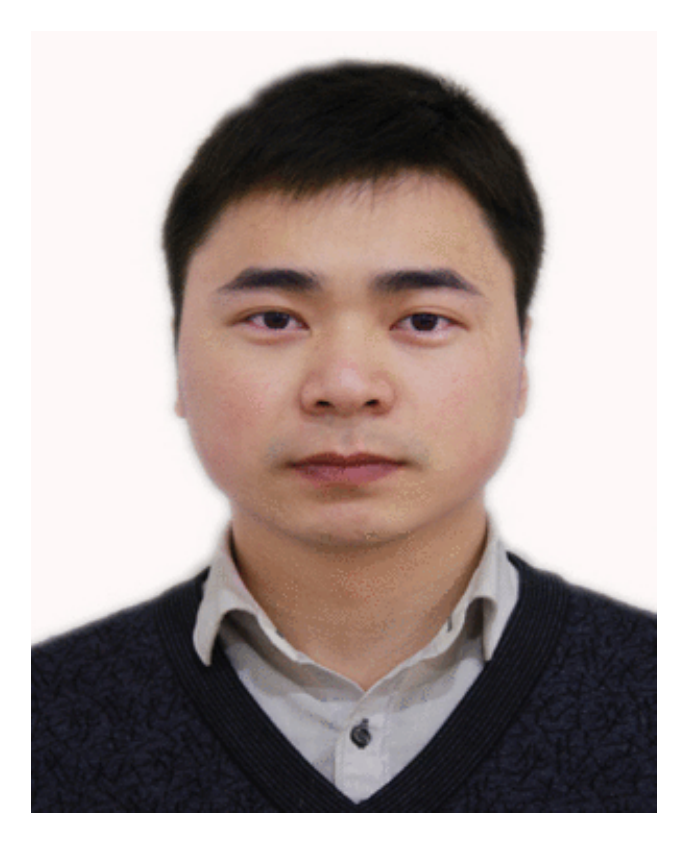}}]{Jingkuan Song}(Senior Member, IEEE) received the Ph.D. degree from The University of Queensland (UQ), St Lucia, QLD, Australia. He then joined Columbia University, as a Postdoc Research Scientist, and the University of Trento, as a Research Fellow. He is currently a Professor with the School of Computer Science and Technology, Tongji University, Shanghai, China. His research interest includes large-scale multimedia retrieval, image/video segmentation and image/video understanding using hashing, graph learning and deep learning techniques. He was the recipient of the Best Paper Award in ICPR (2016, Mexico), Best Student Paper Award at Australian Database Conference (2017, Australia), and Best Paper Honorable Mention Award (2017, Japan). He is also the Guest Editor of IEEE Transactions on Multimedia, and WWWJ and a PC Member of CVPR’18, MM’18, and IJCAI’18.
\end{IEEEbiography}

\begin{IEEEbiography}[{\includegraphics[width=1in,height=1.25in,clip,keepaspectratio]{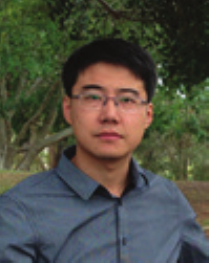}}]{Yang Yang} (Senior Member, IEEE) received the Ph.D. degree in computer science from The University of Queensland, Brisbane, QLD, Australia, in 2012. He was a Research Fellow with the National University of Singapore, Singapore, from 2012 to 2014. He is currently with the University of Electronic Science and Technology of China, Chengdu, China. His current research interests include multimedia content analysis, computer vision, and social media analytics.
\end{IEEEbiography}

\begin{IEEEbiography}[{\includegraphics[width=1in,height=1.25in,clip,keepaspectratio]{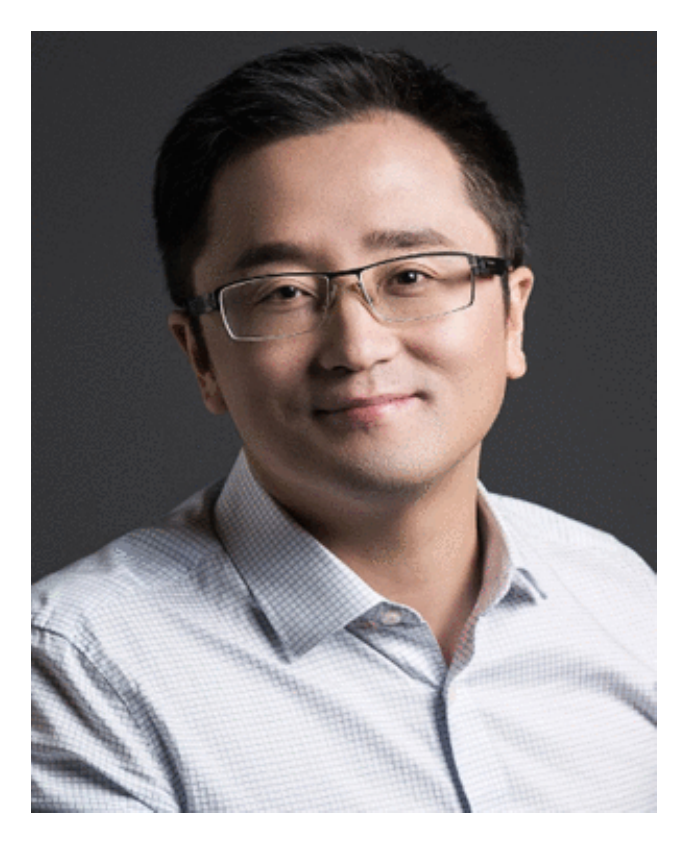}}]{Heng Tao Shen}(Fellow, IEEE) received the B.Sc. (with First-Class Hons.) and the Ph.D. degrees from the Department of Computer Science, National University of Singapore, Singapore, in 2000 and 2004, respectively. He is currently the Dean of the School of Computer Science and Technology, Tongji University, Shanghai, China. His research interests mainly include multimedia search, computer vision, and artificial intelligence. He is a member of Academia Europaea, Fellow of ACM, and OSA.
\end{IEEEbiography}

\vfill

\end{document}